\documentclass[10pt,twocolumn,letterpaper]{article}

\usepackage[pagenumbers]{cvpr} 

\usepackage{makecell} 

\definecolor{cvprblue}{rgb}{0.21,0.49,0.74}
\definecolor{darkred}{rgb}{0.459,0.0,0.08}
\usepackage[pagebackref,breaklinks,colorlinks,allcolors=cvprblue,citecolor=cvprblue,urlcolor=darkred]{hyperref}

\def\paperID{2184} 
\def\confName{CVPR}
\def\confYear{2026}
\usepackage[utf8]{inputenc}
\usepackage[T1]{fontenc}

\usepackage{color,xcolor}
\usepackage{epsfig}
\usepackage{graphicx}

\usepackage{adjustbox}
\usepackage{array}
\usepackage{booktabs}
\usepackage{colortbl}
\usepackage{float,wrapfig}
\usepackage{hhline}
\usepackage{multirow}
\usepackage{subcaption}
\usepackage[font=small]{caption}
\usepackage{makecell}

\usepackage{amsmath,amsfonts,amsthm,amssymb}
\usepackage{bm}
\usepackage{nicefrac}
\usepackage{microtype}

\usepackage{changepage}
\usepackage{extramarks}
\usepackage{fancyhdr}
\usepackage{lastpage}
\usepackage{setspace}
\usepackage{soul}
\usepackage{xspace}
\usepackage{indentfirst}
\usepackage{pifont}
\usepackage{cuted}
\usepackage{wrapfig}
\definecolor{cvprblue}{rgb}{0.21,0.49,0.74}
\usepackage{url}

\usepackage{algorithm, algorithmic}
\usepackage{enumitem}

\usepackage{wasysym}
\usepackage{pifont}
\usepackage{fancyvrb}
\usepackage{fvextra}
\usepackage{epigraph}
\usepackage{mwe}

\usepackage{tabularx}
\usepackage{graphicx}
\usepackage{longtable, ltxtable, booktabs}

\usepackage[textsize=footnotesize, linecolor=magenta, bordercolor=magenta, backgroundcolor=white, textwidth=35pt]{todonotes}
\usepackage{marginnote}
\newcolumntype{L}[1]{>{\raggedright\let\newline\\\arraybackslash\hspace{0pt}}m{#1}}
\newcolumntype{C}[1]{>{\centering\let\newline\\\arraybackslash\hspace{0pt}}m{#1}}
\newcolumntype{R}[1]{>{\raggedleft\let\newline\\\arraybackslash\hspace{0pt}}m{#1}}

\newcommand{\fig}[1]{Figure~\ref{fig:#1}}

\newcommand{\ignorethis}[1]{}

\makeatletter
\DeclareRobustCommand\onedot{\futurelet\@let@token\@onedot}
\def\@onedot{\ifx\@let@token.\else.\null\fi\xspace}

\makeatother

\definecolor{citecolor}{rgb}{34,139,34}
\definecolor{mydarkblue}{rgb}{0,0.08,1}
\definecolor{mydarkgreen}{rgb}{0.02,0.6,0.02}
\definecolor{mydarkred}{rgb}{0.8,0.02,0.02}
\definecolor{mydarkorange}{rgb}{0.40,0.2,0.02}
\definecolor{mydarkgold}{rgb}{0.6, 0.4, 0.05}
\definecolor{mypurple}{RGB}{111,0,255}
\definecolor{myred}{rgb}{1.0,0.0,0.0}
\definecolor{mygold}{rgb}{0.75,0.6,0.12}
\definecolor{mydarkgray}{rgb}{0.66,0.66,0.66}

\newcommand{\myparagraph}[1]{\vspace{0pt}\noindent\textbf{#1}}

\definecolor{mitblue}{rgb}{0.88,0.95,0.96}

\newcommand{\methodshort}{ForeAct\xspace}
\newcommand{\method}{Visual Foresight Planning\xspace}
\newcommand{\planner}{visual foresight planner\xspace}

\usepackage{amsmath,amsfonts,bm}

\def\eqref#1{equation~\ref{#1}}

\def\1{\bm{1}}

\DeclareMathAlphabet{\mathsfit}{\encodingdefault}{\sfdefault}{m}{sl}
\SetMathAlphabet{\mathsfit}{bold}{\encodingdefault}{\sfdefault}{bx}{n}

\ifx\theorem\undefined
\newtheorem{theorem}{Theorem}[section]
\fi

\ifx\example\undefined
\newtheorem{example}{Example}[section]
\fi

\ifx\property\undefined
\newtheorem{property}{Property}
\fi

\ifx\lemma\undefined
\newtheorem{lemma}{Lemma}[section]
\fi

\ifx\proposition\undefined
\newtheorem{proposition}{Proposition}[section]
\fi

\ifx\remark\undefined
\newtheorem{remark}{Remark}
\fi

\ifx\corollary\undefined
\newtheorem{corollary}{Corollary}[section]
\fi

\ifx\definition\undefined
\newtheorem{definition}{Definition}[section]
\fi

\ifx\conjecture\undefined
\newtheorem{conjecture}{Conjecture}[section]
\fi

\ifx\axiom\undefined
\newtheorem{axiom}[theorem]{Axiom}
\fi

\ifx\claim\undefined
\newtheorem{claim}[theorem]{Claim}
\fi

\ifx\assumption\undefined
\newtheorem{assumption}{Assumption}
\fi

\ifx\problem\undefined
\newtheorem{problem}{Problem}[section]
\fi

\ifx\fact\undefined
\newtheorem{fact}{Fact}
\fi

\title{ForeAct: Steering Your VLA with Efficient Visual Foresight Planning}

\makeatletter
\patchcmd{\@maketitle}{\end{center}}{\end{center}\@thanks}{}{}
\makeatother

\author{
Zhuoyang Zhang$^{1}$\thanks{Equal Contribution.} \ \thanks{Part of the work done during an internship at NVIDIA.} \quad
Shang Yang$^{1}$\footnotemark[1] \ \footnotemark[2] \quad
Qinghao Hu$^{1}$ \quad
Luke J. Huang$^{1}$ \quad \\
James Hou$^{1,3}$ \quad
Yufei Sun$^{1}$ \quad
Yao Lu$^{2}$ \quad
Song Han$^{1,2}$ \quad \\
$^{1}$MIT\quad $^{2}$NVIDIA\quad $^{3}$Caltech \\
\href{https://github.com/mit-han-lab/foreact}{\textbf{\textcolor{darkred}{\nolinkurl{https://github.com/mit-han-lab/foreact}}}}
}

\begin{document}
\maketitle
\begin{strip}
\begin{minipage}{\textwidth}
  \centering
  \vspace{-40pt}
  \includegraphics[width=\linewidth]{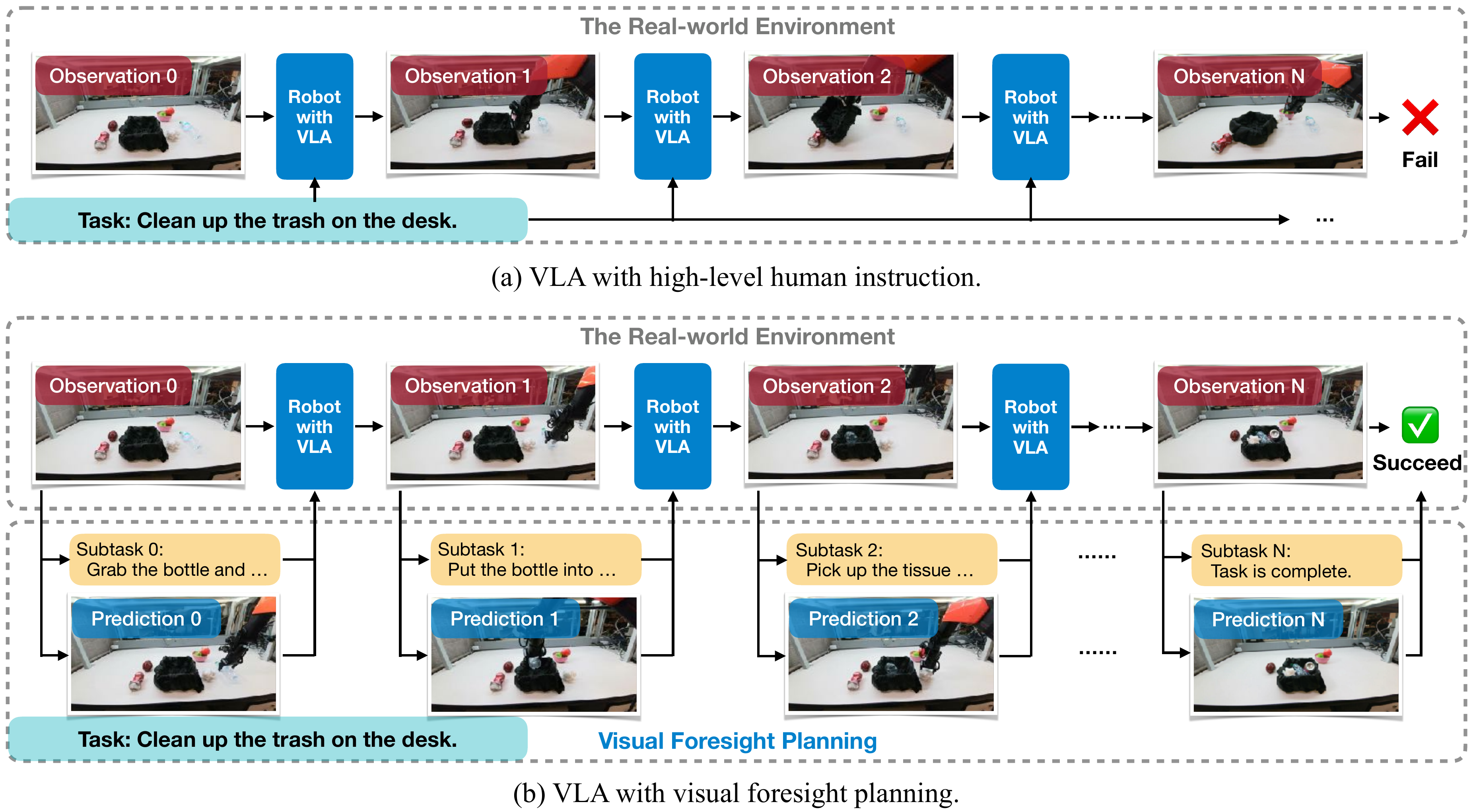}
  \captionsetup{type=figure}
  \vspace{-15pt}
  \caption{\textbf{Illustration of \method (\methodshort).} Instead of relying on a single high-level human instruction, our \planner guides the VLA step-by-step using imagined future observations and subtask descriptions. This design allows the VLA to concentrate on visuo-motor inference rather than high-level semantic reasoning, leading to improved accuracy and better generalization.}
  \label{fig:1_teaser}
\end{minipage}
\end{strip}
\begin{abstract}
Vision-Language-Action (VLA) models convert high-level language instructions into concrete, executable actions, a task that is especially challenging in open-world environments. We present \textit{\method (\methodshort)}, a general and efficient planner that guides a VLA step-by-step using imagined future observations and subtask descriptions. With an imagined future observation, the VLA can focus on visuo-motor inference rather than high-level semantic reasoning, leading to improved accuracy and generalization. Our planner comprises a highly efficient foresight image generation module that predicts a high-quality 640$\times$480 future observation from the current visual input and language instruction within only 0.33s on an H100 GPU, together with a vision-language model that reasons over the task and produces subtask descriptions for both the generator and the VLA. Importantly, state-of-the-art VLAs can integrate our planner seamlessly by simply augmenting their visual inputs, without any architectural modification. The foresight generator is pretrained on over 1 million multi-task, cross-embodiment episodes, enabling it to learn robust embodied dynamics. We evaluate our framework on a benchmark that consists of 11 diverse, multi-step real-world tasks. It achieves an average success rate of 87.4\%, demonstrating a +40.9\% absolute improvement over the $\pi_0$ baseline (46.5\%) and a +30.3\% absolute improvement over $\pi_0$ augmented with textual subtask guidance (57.1\%).
\end{abstract}
\section{Introduction}
\label{sec:intro}
People have been working hard to enhance the physical intelligence of embodied systems, aiming to develop robust and general robots to help humans solve a wide range of tasks in the real world. A truly general robot should be capable of handling complex, diverse, unstructured environments and performing tasks that go far beyond pre-programmed routines. For example, it should be able to manipulate a variety of household objects, assist with daily chores such as cooking, cleaning, or organizing spaces, and adapt its behavior when encountering new tools or unfamiliar layouts.

Recently, vision-language-action (VLA) models~\citep{liu2024rdt, team2024octo, li2024cogact, zhai2025igniting, liao2025genie, cheang2025gr, bu2025univla, team2025gemini, bu2025agibot, shukor2025smolvla, jiang2025galaxea, zitkovich2023rt, kim2024openvla, black2024pi_0, intelligence2025pi_, bjorck2025gr00t, black2023zero} have emerged as a promising direction for developing general-purpose robots. These models learn a direct mapping from visual observations and language instructions to robotic actions in an end-to-end manner. However, most existing VLAs still excel only at simple tasks (e.g., picking up an object and placing it in a designated location), falling short of the broad capabilities we ultimately expect. We posit that one key factor underlying this limitation is the difficulty VLAs face in grounding high-level language instructions into concrete, executable action sequences.

This raises the question: what if we provide VLAs with visual instructions rather than relying solely on language? Instead of merely telling a robot what to do, could we show it how to do it? For instance, when asking a robot to clean a table, supplying an image of the table after cleaning offers richer guidance, potentially simplifying task execution. Yet, in real-world settings, a fully final state may not always be attainable, and even when available, it often remains too abstract, omitting the intermediate steps necessary for execution. For example, cleaning a table including picking up the coke can, the water bottle, and the crumpled tissue, and then disposing of them in the trash bin one by one. This motivates a further inquiry: is it possible to provide robots with step-by-step visual instructions, enabling them to follow a visually guided sequence toward task completion?

To this end, we propose \textit{\method (\methodshort)}, a general and efficient planning framework that instructs the VLA with visual guidance step-by-step. At the core of our approach is an efficient foresight image generator. It acts as a world model that conditions on the current observation and the language instruction to predict future observations. The generator is pretrained on over 1 million multi-task, cross-embodiment episodes curated from open-source datasets~\cite{bu2025agibot, wu2024robomind, jiang2025galaxea, walke2023bridgedata}, enabling it to learn robust embodied dynamics. Complementing this generator, we incorporate a vision-language model~\cite{Qwen3-VL} to reason about complex tasks and infer appropriate subtasks, which are then provided to both the foresight generator and the VLA. These two components constitute our \textit{\methodshort} framework. By supplying imagined future observations, our method substantially reduces the difficulty for VLAs in grounding high-level language instructions into concrete actions. Consequently, the VLA model can focus solely on visuo-motor inference rather than high-level semantic reasoning, leading to improved accuracy and generalization. Notably, state-of-the-art VLAs can seamlessly integrate our visual guidance simply by augmenting their visual inputs, without requiring any architectural modifications.

We highlight the efficiency of our \planner. Running on an H100 GPU, the foresight image generation takes only 0.33s. This demonstrates the practicality of our planner for real-world closed-loop control. To assess its effectiveness, we conduct comprehensive evaluations across 11 diverse real-world tasks. Our method achieves an average success rate of 87.4\%, demonstrating a +40.9\% absolute improvement over the $\pi_0$~\cite{black2024pi_0} baseline (46.5\%) and a +30.3\% absolute improvement over $\pi_0$ augmented with textual subtask guidance (57.1\%).
\section{Related Work}
\label{sec:related_work}

\looseness=-1
\myparagraph{Vision-Language-Action Models.}
Vision-language-action (VLA) models integrate vision, language and action in a single policy, enabling robots to interpret human instructions and act on them. Early approaches~\citep{ahn2022can, driess2023palm, brohan2022rt} demonstrated the potential of multimodal policies for instruction-guided manipulation. More recently, state-of-the-art methods~\citep{liu2024rdt,team2024octo, li2024cogact, zhai2025igniting, liao2025genie, cheang2025gr, bu2025univla, team2025gemini, bu2025agibot, shukor2025smolvla, jiang2025galaxea, zitkovich2023rt, kim2024openvla, black2024pi_0, intelligence2025pi_, bjorck2025gr00t, black2023zero} have advanced by fine-tuning pretrained vision-language models on large-scale robotic datasets, enabling direct mapping from visual-language instructions to robot actions. Representative works including RT-2~\citep{zitkovich2023rt}, OpenVLA~\citep{kim2024openvla}, $\pi_0$~\citep{black2024pi_0}, $\pi_{0.5}$~\citep{intelligence2025pi_}, GR00T-N1~\citep{bjorck2025gr00t}, Gemini Robotics~\citep{team2025gemini} have shown unprecedented capabilities and achieved remarkable performance on real-world robot tasks. However, they still struggle with complex tasks and generalizing to novel scenarios. Our proposed \method method can enhance the VLA's accuracy and generalization.

\begin{figure}[t]
  \centering
  \includegraphics[width=\linewidth]{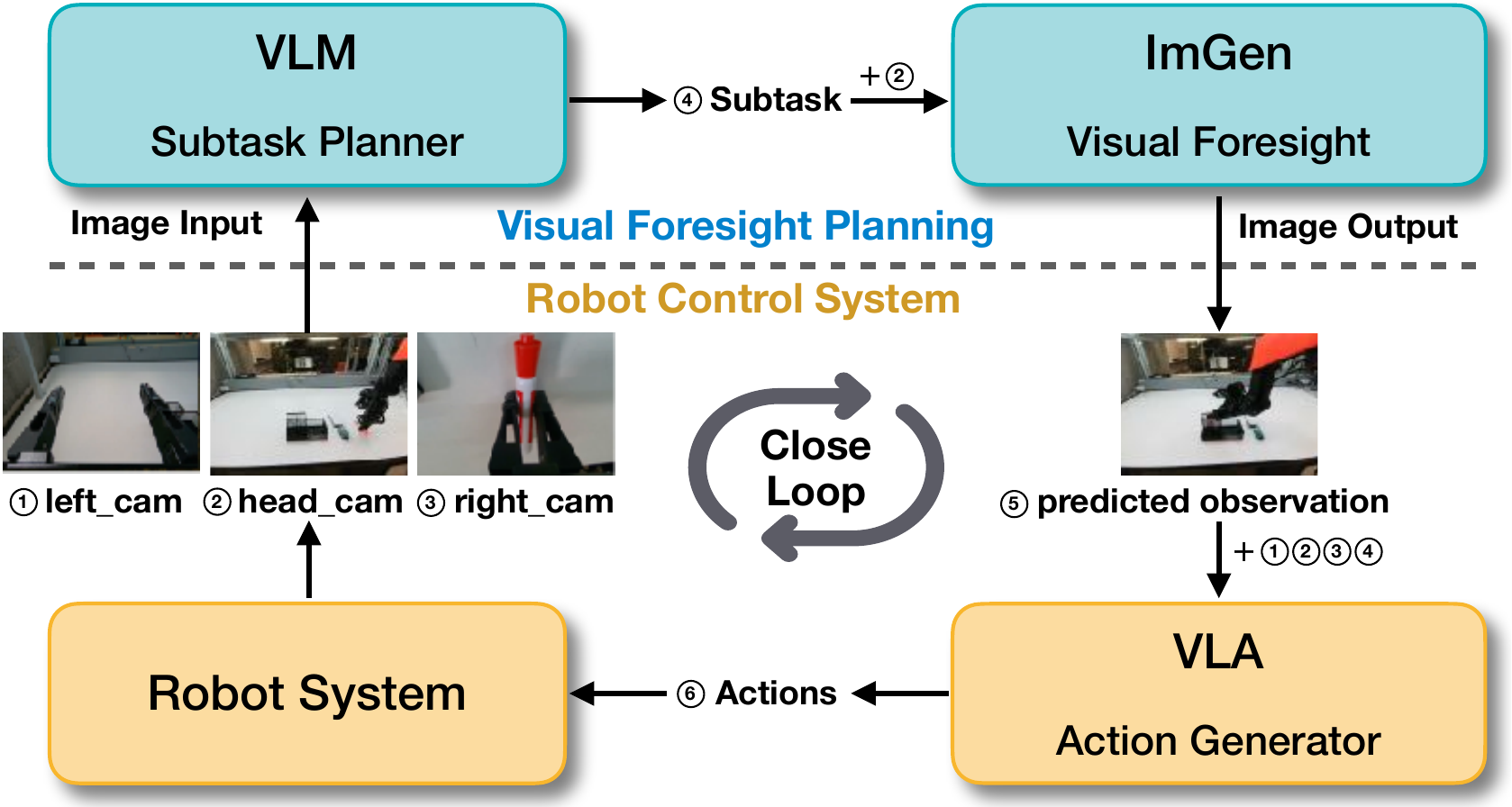}
  \caption{
    \textbf{Overview of our \methodshort framework.} 
    The VLM-based subtask planner takes the robot's head-camera observation and generates a subtask instruction for the Foresight Image Generation (ImGen) module. ImGen then predicts the future observation, which is fed into the VLA model together with the subtask instruction and the robot's three camera views. These modules operate jointly to enable closed-loop control.}
  \label{fig:2_pipeline}
  \vspace{-6pt}
\end{figure}

\myparagraph{Robotic Planning and Reasoning.}
Instead of directly mapping human instructions to robot actions, recent works try to decompose the complex tasks into a sequence of simpler sub-tasks before action execution. One line of research unify planning and control within a single model~\citep{intelligence2025pi_, fang2025robix, zawalski2024robotic, mees2022grounding}. However, such unification introduces significant challenges: (1) VLA models typically employ relatively small vision-language backbones (e.g., 3B parameters) to reduce latency, which limits their capacity for complex reasoning; and (2) fine-tuning these models on robot data often leads to catastrophic forgetting, thereby compromising their general-purpose capabilities. Another line of research adopts a hierarchical framework~\citep{huang2022inner, liang2022code,zeng2022socratic, mees2022grounding, sharma2022skill, singh2022progprompt, zawalski2024robotic, hu2023look, shi2025hi, huang2025thinkact} that delegates planning and reasoning to a separate model. This design mitigates the risk of catastrophic forgetting and enables planning and reasoning to be performed by more capable models. However, it does not fundamentally alleviate the intrinsic challenge faced by VLA models in grounding high-level language instructions into concrete actions. In contrast, our planner provides a more detailed visual guidance, let the VLA focus on visuomotor inference.


\myparagraph{Visual Prediction for Robot Control.}
Integrating visual prediction into the control loop is a key advance to enable lookahead planning. Earlier works~\citep{ebert2018visual, nair2019hierarchical} first investigated the benefit of visual prediction in model-based RL. SuSIE~\citep{black2023zero} generates goal images that guide diffusion policy models. COT-VLA~\citep{zhao2025cot} generates goal images and actions inside a unified model. More recently, with the development of video generation, people start to explore the use of video generation for robot control~\citep{feng2025vidar, ajay2023compositional, du2023learning, ko2023learning, zhou2024robodreamer, bharadhwaj2024gen2act, hu2024video, jang2025dreamgen}. While conceptually appealing, existing video-generation-based methods face key limitations: (1) slow and computationally expensive inference, (2) predominantly open-loop designs that ignore environmental feedback, and (3) incompatibility with state-of-the-art pretrained VLAs. Our method is efficient and general, that can be integrated with the state-of-the-art VLAs in a closed-loop manner.

\section{Method}
\label{sec:method}

In this section, we first present \methodshort with a structured formulation in Section~\ref{sec:visually_grounded_planning}. We then introduce the efficient foresight image generation model, which is the core of our method, in Section~\ref{sec:efficient_foresight_image_generation}. After that, we illustrate how we use VLM for reasoning and monitoring in Section~\ref{sec:vlm_for_reasoning_and_monitoring}. Finally, we show how to seamlessly integrate our method with VLA models in Section~\ref{sec:integration_with_vla} and how we deploy the whole system in Section~\ref{sec:cloud_edge_closed_loop_deployment}.

\subsection{\method (\methodshort)}
\label{sec:visually_grounded_planning}
\myparagraph{VLA.}
Modern state-of-the-art VLAs are generally implemented as language-conditioned models, which aim to learn the conditional distribution
\[
\pi(\boldsymbol{A}_t \mid \boldsymbol{I}_t, \boldsymbol{q}_t, l)
\]
where $\boldsymbol{A}_t = [a_t, a_{t+1}, \cdots, a_{t+H-1}]$ is a chunk of actions of length H, $\boldsymbol{I}_t = [I_t^1, I_t^2, \cdots, I_t^K]$ is the current visual observation from multiple cameras, $\boldsymbol{q}_t$ is the proprioceptive states from the robot, and $l$ is the language instruction.

\myparagraph{VLA with \methodshort.} Our key insight is to introduce a \planner to enhance the VLA model. With \methodshort, the conditional distribution is reformulated as follows:
\begin{equation*}
    \begin{aligned}
    \pi(\boldsymbol{A}_t \mid \boldsymbol{I}_t, \boldsymbol{q}_t, l)
    &= \pi_l(\boldsymbol{A}_t \mid [\boldsymbol{I}_t, \boldsymbol{G}_t], \boldsymbol{q}_t, l_t) \pi_h(\boldsymbol{G}_t, l_t \mid \boldsymbol{I}_t, l)
    \end{aligned}
\end{equation*}
$\pi_h$ is our \planner, it provides the predicted future observation $\boldsymbol{G}_t$ and subtask description $l_t$ conditioned on the current observation and the overall task description. $\pi_l$ is the VLA model, it takes the predicted future observation $\boldsymbol{G}_t$ as an additional visual input and the subtask description $l_t$ as the language condition to generate actions. Specifically, $\pi_h$ is implemented with two components: 
\begin{equation*}
    \begin{aligned}
    \pi_h(\boldsymbol{G}_t, l_t \mid \boldsymbol{I}_t, l)
    &= \pi_g(\boldsymbol{G}_t \mid \boldsymbol{I}_t, l_t) \pi_v(l_t \mid \boldsymbol{I}_t, l)
    \end{aligned}
\end{equation*}
At the core is the foresight image generation model $\pi_g$, which grounds the high-level language instruction into a concrete predicted future observation. Complementing this, $\pi_v$ is a VLM to reason over the complex task and infer the subtask. \fig{2_pipeline} illustrates the overall pipeline of our \methodshort framework.

\subsection{Efficient Foresight Image Generation}
\label{sec:efficient_foresight_image_generation}
Our core idea is to predict an imagined future observation that clearly guides the VLA model, rather than relying on solely language instructions. We design the foresight image generation model to closely approximate the actual future observation. To preserve generalization across heterogeneous robotic platforms, the model is conditioned solely on visual observations and language instructions, excluding other modalities such as robot proprioceptive states. For better adaptability and efficiency, we generate the foresight image only for the head camera, which captures the global scene and thus provides the most informative context.

\begin{figure}[t]
  \centering
  \includegraphics[width=\linewidth]{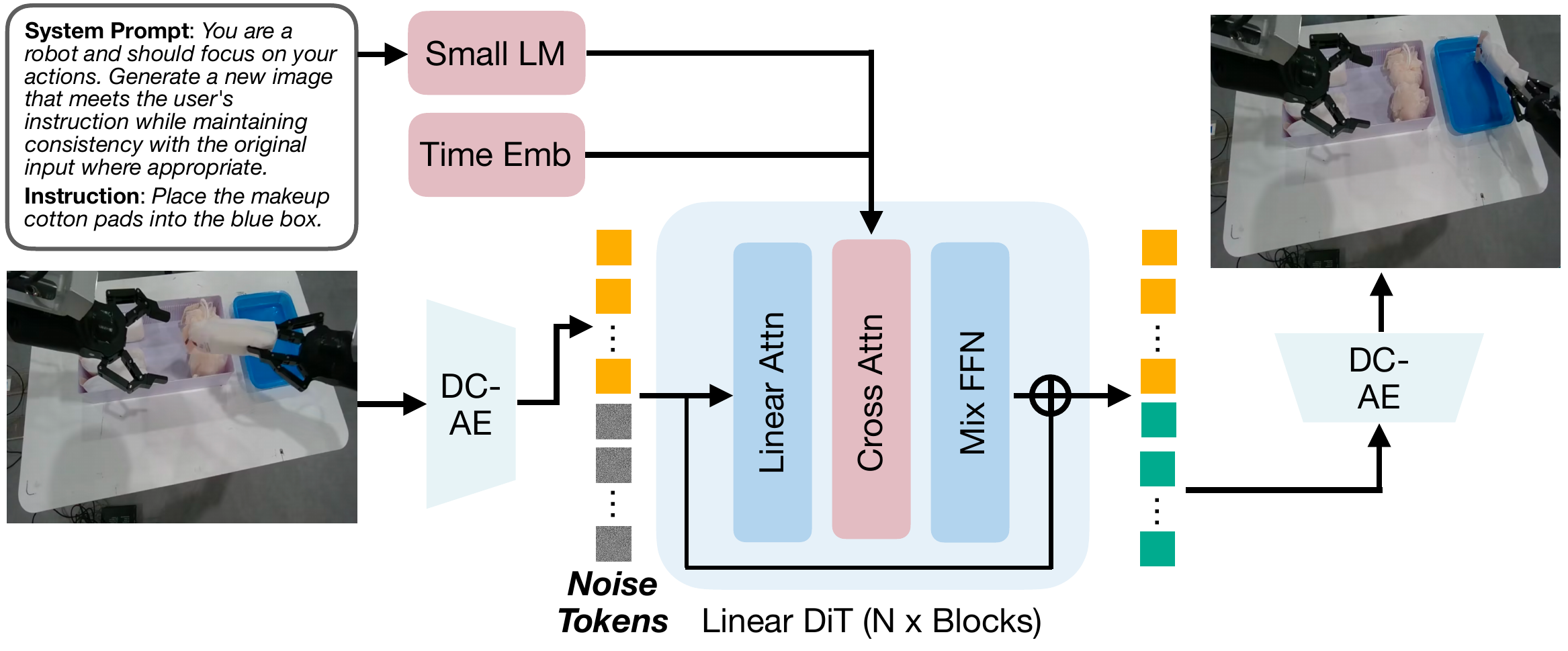}
  \caption{\textbf{Foresight image generation model.} The current observation is first encoded into compact visual tokens and concatenated with a noise input, then fed into the efficient linear DiT model to generate the predicted visual tokens, which are subsequently decoded into the future observation. In addition, the instruction and a specially designed system prompt are incorporated to guide the model's attention toward the robot's actions.}
  \label{fig:3_foresight_image_generation}
  \vspace{-10pt}
\end{figure}

\myparagraph{Model Architecture.} Modern robots typically employ high-resolution cameras (e.g., 640$\times$480 or even 1280$\times$720) to capture detailed visual information. Consequently, our foresight image generation model must also support high-resolution outputs to preserve fine details and subtle movements. Moreover, the generation must be efficient to be engaged in the real-world closed-loop control. Inspired by recent advances in efficient image generation, we adopt the architectural design of the state-of-the-art model SANA~\cite{xie2024sana}. SANA employs a 32$\times$ deep compression autoencoder~\cite{chen2024deep} to encode images into a compact latent space, substantially reducing the number of latent tokens. It also utilizes a linear DiT architecture to enable efficient attention computation at high resolutions. However, SANA is originally designed for text-to-image generation and thus cannot directly process image conditions. To address this limitation, we concatenate the conditioned image with the noise input, thereby transforming the denoising process into a conditional denoising task. The detailed model architecture of our foresight image generation model is illustrated in Figure \ref{fig:3_foresight_image_generation}.

\begin{figure}[t]
    \centering
    \includegraphics[width=0.95\linewidth]{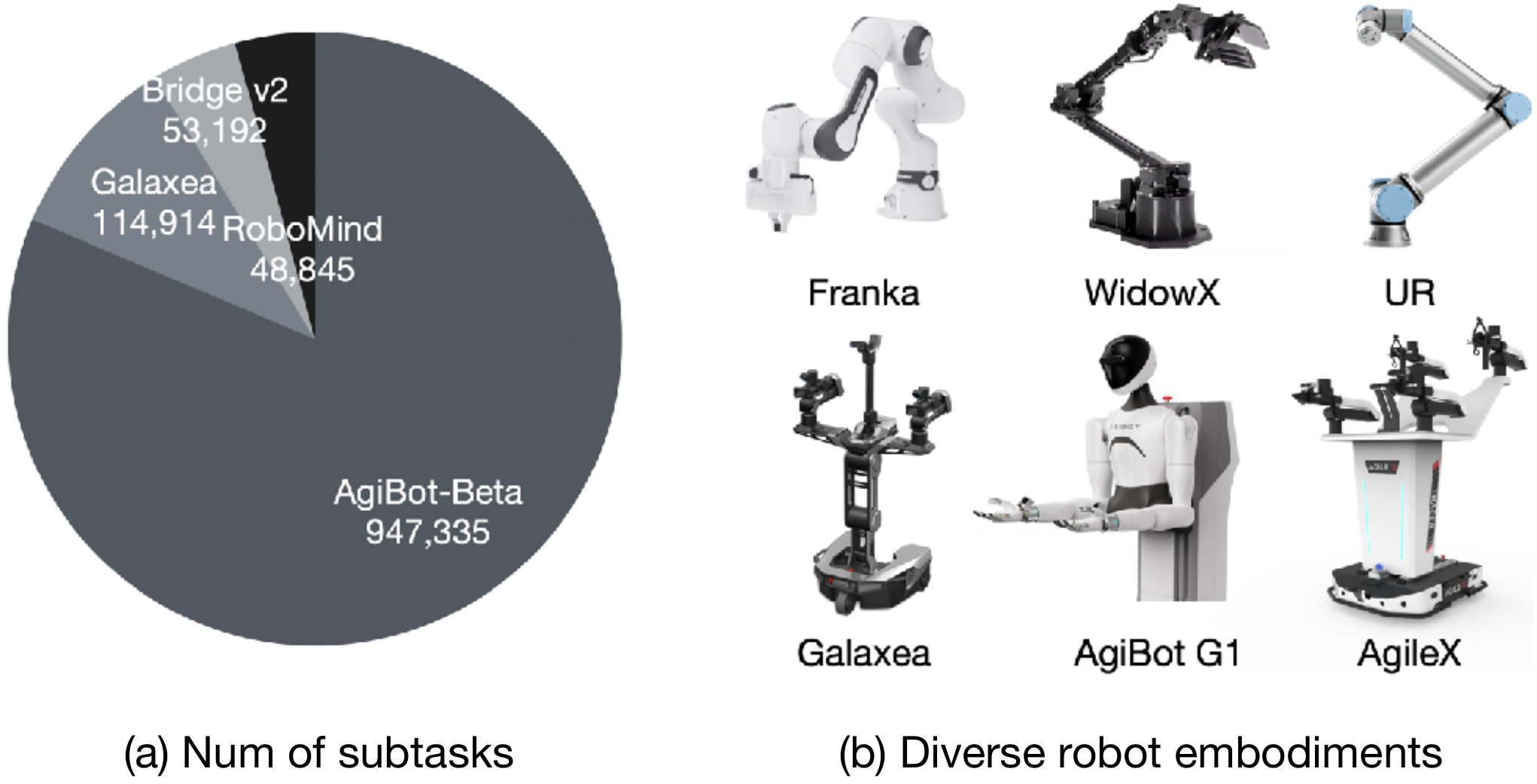}
    \caption{(a) \textbf{Number of subtasks from each dataset.} Our data comes from a wide range of sources, and preprocessing yields a total of 1.16 million subtasks. (b) \textbf{Diverse robot embodiments in the pre-training dataset.} The collected dataset covers a wide range of robot embodiments.}
    \label{fig:4_pretrain_data}
    \vspace{-6pt}
  \end{figure}

\myparagraph{Pre-training Dataset.} We collect a large and diverse open-source cross-embodiment dataset for pre-training, allowing the model to learn generalizable embodied dynamics through foresight image generation. Specifically, we collect data from the AgiBot-World Colosseo~\cite{bu2025agibot}, RoboMind~\cite{wu2024robomind}, Galaxea Open-World~\cite{jiang2025galaxea}, and Bridge~\cite{walke2023bridgedata} datasets. We exclude datasets such as Open-X-Embodiment~\cite{o2024open} and DROID~\cite{khazatsky2024droid} due to their relatively low camera resolution. Our collected dataset covers a wide range of robot embodiments as shown in the right panel of Figure \ref{fig:4_pretrain_data}.

\myparagraph{Data Preprocessing.}
After collecting the datasets, we preprocess the data to suit our requirements. Tasks in AgiBot-World Colosseo, RoboMind, and Galaxea Open-World are typically complex and long-horizon, so we decompose them into subtasks. They already provide detailed subtask segmentations, which we directly adopt for task splitting. In contrast, the Bridge dataset contains relatively simple tasks, so we use the provided task descriptions directly without further subdivision. After preprocessing, the number of subtasks from each dataset is shown in the left panel of Figure \ref{fig:4_pretrain_data}, with a total number of 1.16 million subtasks.

Within each subtask, we sample condition frames at 1-second intervals. For each condition frame, we choose a future frame that comes half a subtask's length later in time. We set this offset to half the subtask length because it gives frame pairs that clearly show meaningful action changes. The paired language instruction is the description of the robot's action of the subtask. After sampling, we get a total of approximately 10 million data pairs for pre-training.

\myparagraph{Training Details.}
Following SANA, we adopt flow matching as our training objective. All images are resized to 640×480 during training. We initialize our model with SANA-1.6B-512px's pre-trained weights to leverage its strong image-generation priors. The global batch size is set to 512, and training is conducted for 800K steps on 64 H100 GPUs. We leverage a constant learning rate of 5e-5 with 5K steps warmup.

\subsection{VLM for Reasoning and Monitoring}
\label{sec:vlm_for_reasoning_and_monitoring}
In our \methodshort framework, the VLM operates in a \textit{reason-execute-monitor} cycle. Given the overall task description and the current observation, it first reasons by generating an immediate, actionable subtask. After the VLA executes the actions, the VLM monitors by assessing the updated state, tracking progress, and re-planning the next subtask if the current subtask is completed. This hierarchy enables the VLA model to dynamically recover from execution failures and adapt to environmental uncertainties, yielding robust and scalable performance across complex real-world tasks. 

\subsection{Adapt to VLA Models}
\label{sec:integration_with_vla}
When integrating \methodshort, the VLA model requires no architectural modifications as we only augment the visual input. During VLA fine-tuning, we concatenate the current observation with a future observation to form the visual input. During inference, the generated foresight image is appended to the current observation to serve as the visual input, while the generated subtask instruction serves as the language input to the fine-tuned VLA model. This design requires minimal adjustments, thereby enabling seamless integration of our \methodshort with existing VLA models. 

\subsection{Cloud-Edge Closed-Loop Deployment}
\label{sec:cloud_edge_closed_loop_deployment}
To achieve responsive, closed-loop control for real-world robotics, we design a hierarchical cloud-edge deployment pipeline. The workflow begins with the edge host responsible for reactive local control, which captures and sends the observation to the remote cloud server. This server, specialized for reasoning and foresight generation (hosting the VLM and image generator with vLLM~\cite{vllm}), performs high-level planning and distills the result into a dual-guidance packet (textual $l_t$, visual $\boldsymbol{G}_t$). This packet is then returned to the edge host, where the local VLA policy performs rapid inference to dispatch the final action. In our setting, we deploy the \planner on an H100 GPU, and the VLA on a RTX 5090 GPU.
\section{Experiments}
\label{sec:experiments}
\subsection{Setups}
\label{sec:setups}


\myparagraph{Robot System.}
We collect our real-world data on the Galaxea R1 Lite~\cite{R1LiteDoc}, a dual-arm mobile manipulator equipped with 23 degrees of freedom (6-DoF chassis, 3-DoF torso, and two 7-DoF arms). The platform also features a comprehensive multi-view perception suite, including a head-mounted stereo camera and two wrist-mounted monocular depth cameras.

\myparagraph{Comprehensive Real-World Dataset.}
We collect a comprehensive and challenging real-world dataset consisting of 11 tasks that simulate activities in \textit{Kitchen}, \textit{Workspace}, and \textit{Factory} with varying levels of difficulty.

\begin{figure}[t]
  \centering
  \includegraphics[width=\linewidth]{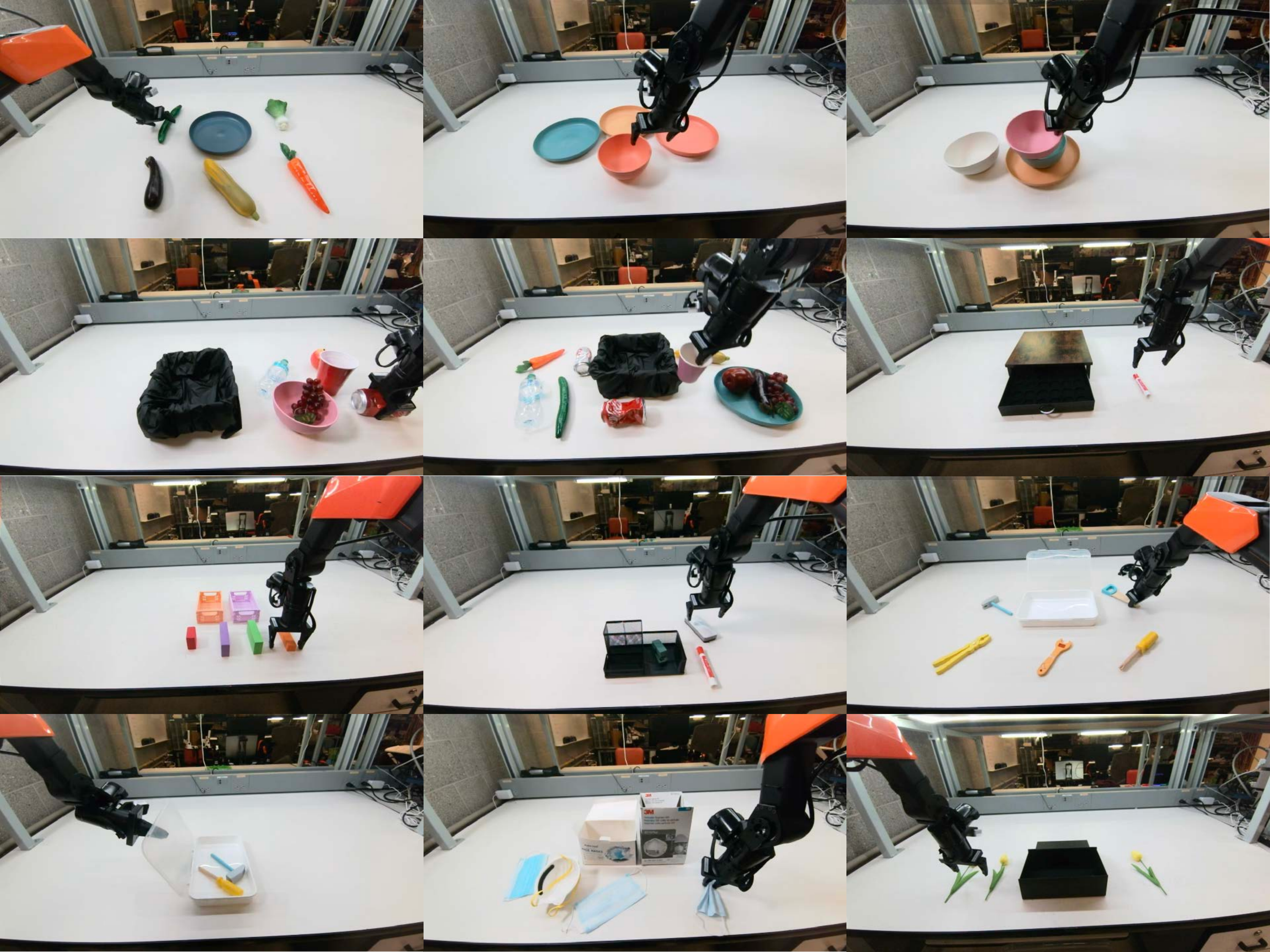}
  \caption{\textbf{Examples of tasks in our real-world dataset.}}
  \label{fig:5_robot_data}
\end{figure} 

\begin{itemize}
\item \textbf{Kitchen:} Tasks involve diverse object manipulation in kitchen environments, ranging from simple conditional placements (\textit{e.g., pick one of five vegetables on a plate}) to multi-stage sequential operations (\textit{e.g., stack three bowls on a plate}) and complex dual-arm sorting (\textit{e.g., classify tabletop waste into bins using single or dual arms}).
\item \textbf{Workspace:} Tasks emphasize typical office organization routines, scaling from basic interactions (\textit{e.g}., \textit{put a pen in a drawer}) to complex, multi-step arrangements involving multiple objects (\textit{e.g., organize a study desk} by placing pens, staplers, and tape into their designated containers).
\item \textbf{Factory:} Tasks simulate industrial kitting and assembly procedures, including basic tool retrieval (\textit{e.g., place a specified tool into a box}), item packing (\textit{e.g., pack flowers into a box}) and instance-level sorting (\textit{e.g., separate different masks into different boxes}).
\end{itemize}

We provide example tasks of our real-world dataset in Figure~\ref{fig:5_robot_data}. 
In total, we collect 420 episodes. Each episode is then decomposed into subtasks, resulting in 2,312 subtask-level episodes. We randomly sample 50 episodes to construct the in-distribution test set, the remaining 370 episodes are used for finetuning the foresight image generation model and the VLA model. 

\subsection{Efficient Foresight Image Generation}
\label{sec:foresight_image_generation}
We have already pretrained the foresight image generation model on a comprehensive and diverse dataset. For deployment in our specific setting, we perform lightweight finetuning to adapt to our specific robot type and the new environment. We finetune the foresight image generation model on our collected real-world data. We use a batch size of 32 and run for 5 epochs. We leverage a constant learning rate of 1e-5 with no warmup.

\myparagraph{Effectiveness of Large-scale Pre-training.}
To demonstrate the effectiveness of the pretraining, we also conduct an experiment in which the foresight image generation model is trained using only our self-collected data without robot data pretraining. For fair comparison, the from scratch training follows the same settings as the finetuning.

We evaluate the generated images under both in-distribution and out-of-distribution test settings. The in-distribution setting corresponds to tasks and environments that match those seen during training, although the object layouts may differ. Since layout variations are also present in the training data, we consider these cases as in-distribution. The out-of-distribution setting involves changes to either the task or the environment. For example, the robot may be required to manipulate a different object within the same environment, such as picking the corn in the packing flower environment. For the in-distribution evaluation, we sample 50 observations from the test split of our dataset; for the out-of-distribution evaluation, we design and collect 50 new observations.

We assess generation performance from two perspectives: (1) image fidelity, which measures whether the generated image accurately follows the subtask instruction and lies on the correct path toward task completion; and (2) image quality, which evaluates whether the generated image preserves the detailed information of the original image without introducing significant distortions. Both criteria are scored via human evaluation, where each image is scored as 1 if it satisfies the requirement and 0 otherwise. As shown in Table~\ref{tab:image_generation}, the foresight image generation model with pretraining achieves substantially better performance than the model trained from scratch. Without pretraining, the model completely fails on out-of-distribution tasks and obtains only very low scores on in-distribution tasks, indicating that it learns virtually no generalizable capability. In contrast, our pretrained model consistently attains high scores on both in-distribution and out-of-distribution tasks, demonstrating the effectiveness of the pretraining.

\begin{table}[t]
    \centering
    \small
    \caption{\textbf{Quantitative results of foresight image generation.} The scores are averaged over 50 images for both in-distribution and out-of-distribution tasks.}
    \setlength{\tabcolsep}{3pt}
    \begin{tabular}{lcccc}
    \toprule
    & \multicolumn{2}{c}{\textbf{In distribution}} 
    & \multicolumn{2}{c}{\textbf{Out of distribution}} \\
    \cmidrule(lr){2-3} \cmidrule(lr){4-5}
    \textbf{Type} & \textbf{Fidelity} & \textbf{Quality} & \textbf{Fidelity} & \textbf{Quality} \\
    \midrule
    w/o pretraining  & 0.18 & 0.24 & 0.00 & 0.00 \\
    w/ pretraining & 1.00 & 1.00 & 0.88 & 0.96 \\
    \bottomrule
    \end{tabular}
    \label{tab:image_generation}
\end{table}

\begin{figure}[t]
  \centering
  \includegraphics[width=\linewidth]{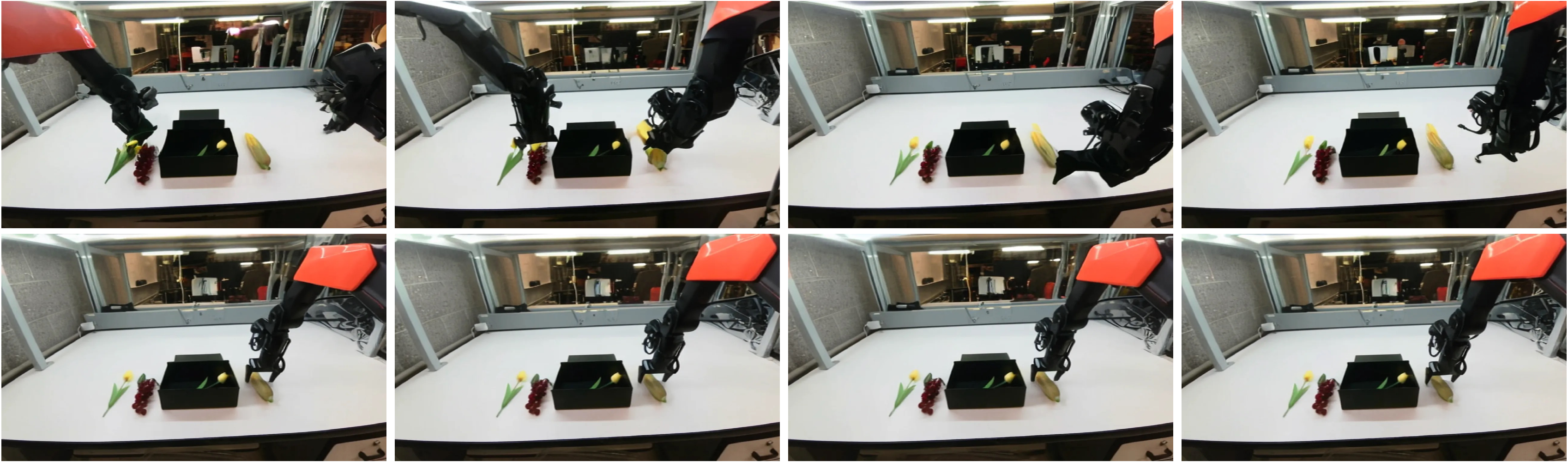}
  \caption{\textbf{Qualitative results of foresight image generation.} The first row shows the model without pretraining and the second row shows the model with pretraining. The task is to pick up the corn. We generate four images for each model with different random seeds.}
  \label{fig:6_qualitative_results_foresight_image_generation}
  \vspace{-10pt}
\end{figure}

We also provide a qualitative example for an out-of-distribution task in Figure~\ref{fig:6_qualitative_results_foresight_image_generation}. The task is to pick up the corn. We generate four images for each model with different random seeds. The first row shows the model without pretraining, which can not generate the correct predicted image. The second row shows the model with pretraining, which generates the correct predicted image across all the four images, showing the effectiveness of the pretraining.

\begin{figure}[t]
    \centering
    \includegraphics[width=\linewidth]{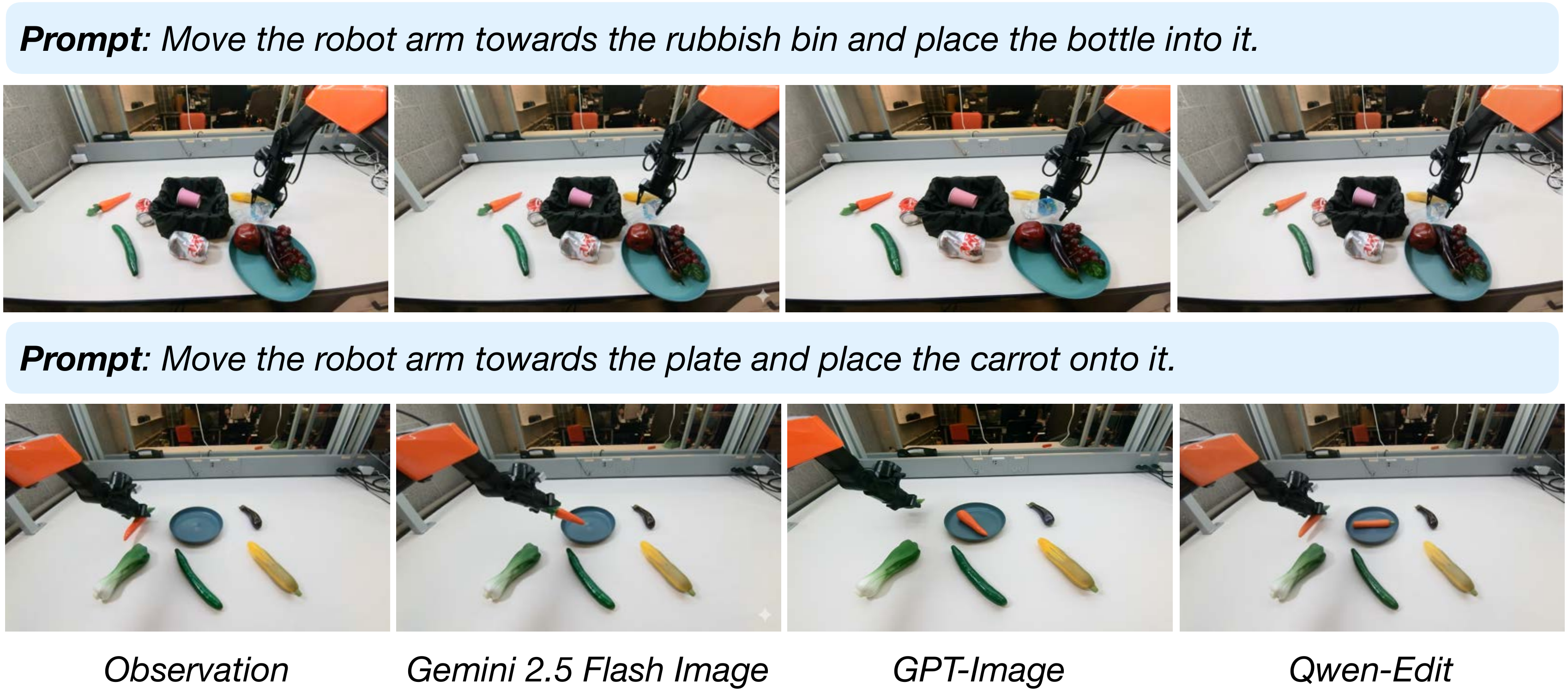}
    \caption{\textbf{Qualitative results of image editing models.} State-of-the-art image-editing models lack an understanding of the underlying real-world physics and object dynamics. They can not generate the correct predicted image with correct robot action.}
    \label{fig:7_qualitative_results_image_editing}
  \end{figure}

\myparagraph{Discussion with Image Editing Models.} Our foresight image generation model conditions on the current observation (image) and a subtask instruction (text) to produce the predicted next-step image. This workflow is closely related to that of image-editing models, which raises the question: Can state-of-the-art image-editing models handle our task? To investigate this, we provide a qualitative comparison in Figure~\ref{fig:7_qualitative_results_image_editing} and evaluate several state-of-the-art image-editing models, including Gemini 2.5 Flash Image~\cite{comanici2025gemini}, GPT-Image~\cite{OpenAI_GPTImage1_Doc}, and Qwen-Edit~\cite{wu2025qwenimagetechnicalreport}. As shown in the first example, all three models make almost no meaningful modifications to the scene. In the second example, although the first model slightly adjusts the arm, it produces an unnatural grasp of the carrot, while the other two models directly hallucinate a carrot on the plate instead of manipulating the existing one. These failure cases indicate that state-of-the-art image-editing models lack an understanding of the underlying real-world physics and object dynamics, thus can not be used to handle our task. Moreover, due to their large parameter sizes, these models incur substantial inference latency, making them impractical for use in closed-loop control.
\begin{figure}[t]
    \centering
    \includegraphics[width=\linewidth]{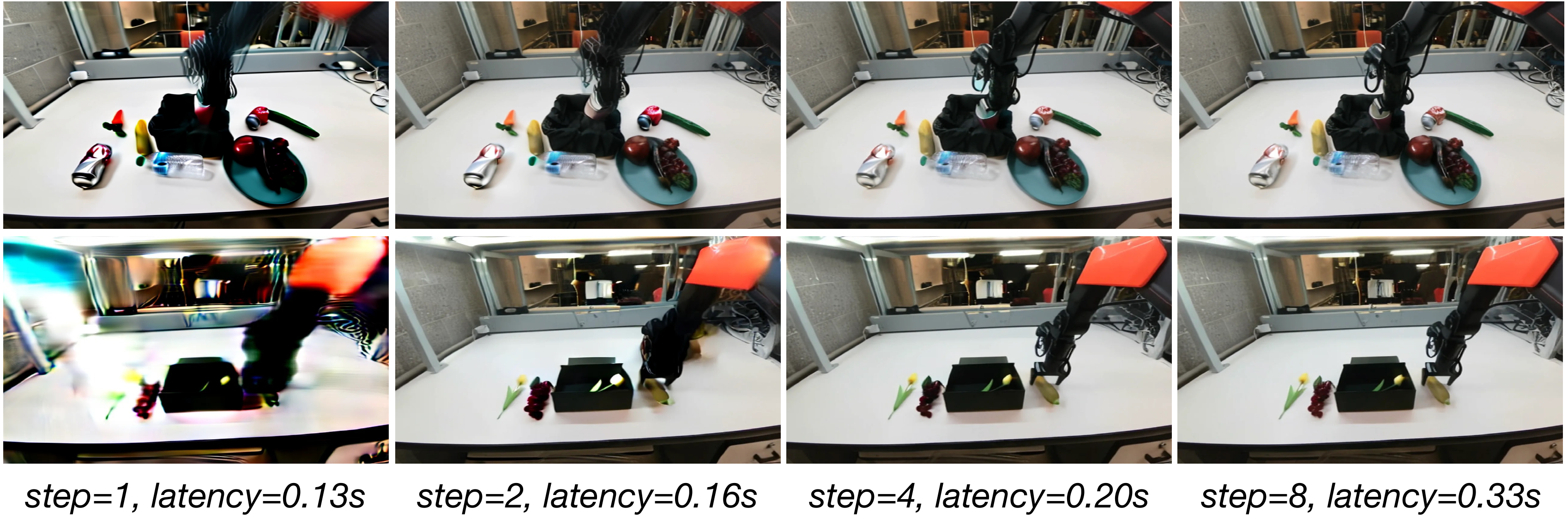}
    \caption{\textbf{Latency measurement of foresight image generation.} Our foresight image generation model is able to generate a high-quality 640$\times$480 imagined future observation in 0.33 seconds on an H100 GPU.}
    \label{fig:9_latency_foresight_image_generation}
    \vspace{-10pt}
  \end{figure}
\begin{figure*}[t]
  \centering
  \includegraphics[width=\linewidth]{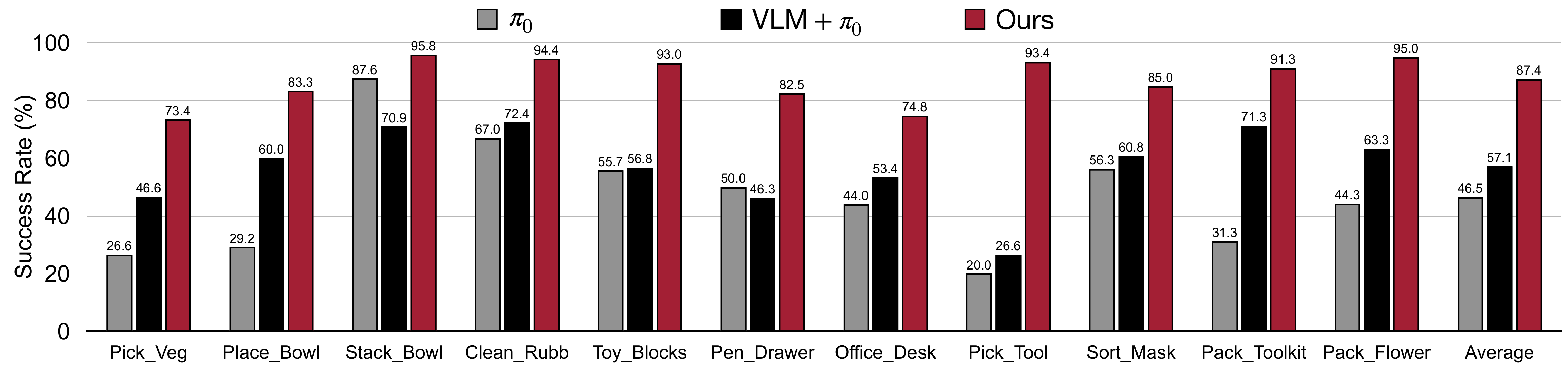}
  \caption{
    \textbf{Real-world evaluation of \methodshort framework}. Our system consistently outperforms baselines without visual guidance across all benchmarking tasks. It achieves an average success rate of 87.4\%, representing a +30.3\% improvement over the VLM-augmented baseline (VLM + $\pi_0$) and a +40.9\% improvement over the vanilla $\pi_0$ baseline. Furthermore, it maintains an average success rate of over 70\% across all benchmark tasks, demonstrating the robustness of our proposed method. 
  }
\label{fig:8_main_results}
\end{figure*}

\myparagraph{Latency Measurement.} We evaluate the latency of our foresight image generation model on an H100 GPU, where the planner is deployed. Because latency is critical for closed-loop control, minimizing generation time is essential. A straightforward approach is to reduce the number of denoising steps. As shown in Figure~\ref{fig:9_latency_foresight_image_generation}, extremely small step counts (e.g., 1-2) lead to severe artifacts, whereas using 8 steps yields visually reliable predictions. Increasing the step count beyond 8 provides negligible quality improvements. Therefore, in our real-world experiments, we set the number of steps to 8, resulting in a latency of only 0.33 s. This demonstrates the efficiency of our foresight image generator and highlights its practicality for real-world closed-loop control. Although we have not yet incorporated acceleration techniques such as distillation or quantization, we believe further reductions in latency are possible and leave this exploration for future work.

\subsection{Real-World Benchmark Evaluation}
\label{sec:end_to_end_evaluation}


\myparagraph{Baselines.} We set up end-to-end real-world evaluation on the Galaxea R1 Lite platform~\cite{R1LiteDoc} with the $\pi_0$~\cite{black2024pi_0} architecture and finetuning it on our self-collected data for real-world adaptation. We compare with the following systems to illustrate the effectiveness of \methodshort:

\begin{itemize}
  \item \emph{$\pi_0$}: A 3.3B VLA model pretrained on over 10k hours of robot data. We finetune it on our self-collected data on Galaxea R1 Lite platform for real-world adaptation.
  \item  \emph{VLM + $\pi_0$}: $\pi_0$ model augmented with Qwen-3-VL-8B-Instruct~\cite{Qwen3-VL} model for subtask planning and monitoring. We finetune the $\pi_0$ model on the subtask-level episodes of our self-collected data, in accordance with the VLM planning granularity.
  \item  \emph{Ours}: $\pi_0$ model enhanced with our \methodshort framework. In our \planner, we also employ Qwen-3-VL-8B-Instruct as the VLM for subtask planning and monitoring. We fine-tune the $\pi_{0}$ model on the subtask-level episodes from our self-collected dataset, augmenting the visual input with a future observation. Specifically, we use the final frame of each subtask episode as the future observation.
\end{itemize}


\myparagraph{Metrics.} We evaluate the performance of the systems across 11 diverse tasks. Please refer to Sec.~\ref{supp:task_desp} for detailed description of all our real-world tasks with visual examples.
For each task, we break down the task into "atomic actions" at the granularity of "approach and grasp" and "move and place". Then, we score the systems success rate of the whole task as the ratio of the number of successful "atomic actions" to the total number of "atomic actions". Meanwhile, if the robot performs any undesired action, such as picking up any object that is not targeted in the task, we deduct the score by the value of a single atomic action. If any atomic action cannot be completed within 5 trials, or the robot ruins the whole task setting through undesired actions, we consider the uncompleted "atomic actions" in the task as failed. For each task, we repeat the evaluation with at least 5 different initial settings (e.g., different object layouts) and report the average score of each model. 

\myparagraph{Results.} In \fig{8_main_results}, we demonstrate the evaluation results on our real-world benchmark. Across all the 11 tasks, our system achieves the highest success rate, with a +30.3\% improvement over the VLM + $\pi_0$ baseline and a +40.9\% improvement over the vanilla $\pi_0$ baseline, demonstrating the effectiveness of our \methodshort. 


Particularly, for the tasks such as Pack\_Flower and Toy\_Blocks, where the robot need to distinguish different objects and take different actions according to the object type, our system surpasses the baseline systems by a great margin. We attribute this improvement to our visual foresight guidance, which provides the robot with the additional goal state information rather than brief text instructions. As a result, the VLA model does not bother to reason which is the aimed object and what is the intended action, and the visual guidance helps the VLA model to learn generalizable visuo-motor skills. 


Additionally, to demonstrate the generality and scalability of our framework, we also evaluate it with the stronger $\pi_{0.5}$ backbone. As detailed in Table~\ref{tab:pi05_realworld}, our method consistently outperforms the vanilla $\pi_{0.5}$ baseline. Overall, the average success rate increases from 70.3\% to 88.2\%, confirming that our visual foresight planning provides essential enhancement for general VLA models.

\begin{table}[t]
\centering
\begingroup
\small
\caption{\textbf{Real-world evaluation with $\pi_{0.5}$.} On a representative set of real-world tasks, our method boosts the average success rate of the highly capable $\pi_{0.5}$ model from 70.3\% to 88.2\%. }
\label{tab:pi05_realworld}
\begin{tabular}{ccc | ccc}
\toprule
\textbf{Task Name} & {$\mathbf{\pi_{0.5}}$} & \textbf{Ours}  & \textbf{Task Name} & {$\mathbf{\pi_{0.5}}$} & \textbf{Ours} \\ 
\midrule
Pick\_Veg & 60.0 & 86.6 & Office\_Desk & 76.0 & 85.4 \\
Place\_Bowl & 75.0 & 83.3 & Pick\_Tool & 50.0 & 96.7 \\
Pen\_Drawer & 68.8 & 81.3 & Pack\_Flower & 91.8 & 95.8 \\
\bottomrule
\end{tabular}
\endgroup
\end{table}

\subsection{Simulation Benchmark Evaluation}
\label{sec:sim_exp}

We further validate our method with the LIBERO simulation benchmark~\cite{liu2023libero} with $\pi_{0.5}$~\cite{intelligence2025pi_} model. We evaluated the performance across four standard task suites: Spatial, Object, Goal, and Long-horizon (LIBERO\_10). The success rate is averaged over 500 trials for each suite. As shown in Table~\ref{tab:libero}, our method consistently improves the performance upon $\pi_{0.5}$ baseline that already achieves near-perfect success rates, raising the average success rate from 96.8\% to 97.5\%.

\begin{table}[t]
    \centering
    \setlength{\tabcolsep}{5pt}
    \caption{\textbf{Simulation evaluation on LIBERO.} We evaluate the success rate (\%) of our method integrated with $\pi_{0.5}$ against state-of-the-art VLA baselines including vanilla $\pi_{0.5}$.}
    \vspace{-5pt}
    \label{tab:libero}
    \scalebox{0.88}{
    \begin{tabular}{c|cccc|c}
        \toprule
        \textbf{Method}  & \textbf{Spatial} & \textbf{Object} & \textbf{Goal} & \textbf{Long} & \textbf{Average} \\
        \midrule
        OpenVLA~\cite{kim2024openvla} & 84.7 & 88.4 & 79.2 & 53.7 & 76.5 \\
        CoT-VLA~\cite{zhao2025cot} & 87.5 & 91.6 & 87.6 & 69.0 & 83.9 \\
        $\pi_{0}$~\cite{black2024pi_0} & 96.8 & 98.8 & 95.8 & 85.2 & 94.2 \\
        $\pi_{0.5}$~\cite{intelligence2025pi_} & 97.3 & 98.8 & 96.9 & 94.2 & 96.8 \\
        CogVLA~\cite{li2025cogvla} & \textbf{98.6} & 98.8 & 96.6 & 95.4 & 97.4 \\
        Ours (w/ $\pi_{0.5}$) & 97.3 & \textbf{99.8 }& \textbf{97.3 }& \textbf{95.4} & \textbf{97.5 }\\
        \bottomrule
    \end{tabular}
    }
\end{table}

\subsection{Generalization to OOD Tasks}

\begin{figure}[t]
  \centering
  \includegraphics[width=0.95\linewidth]{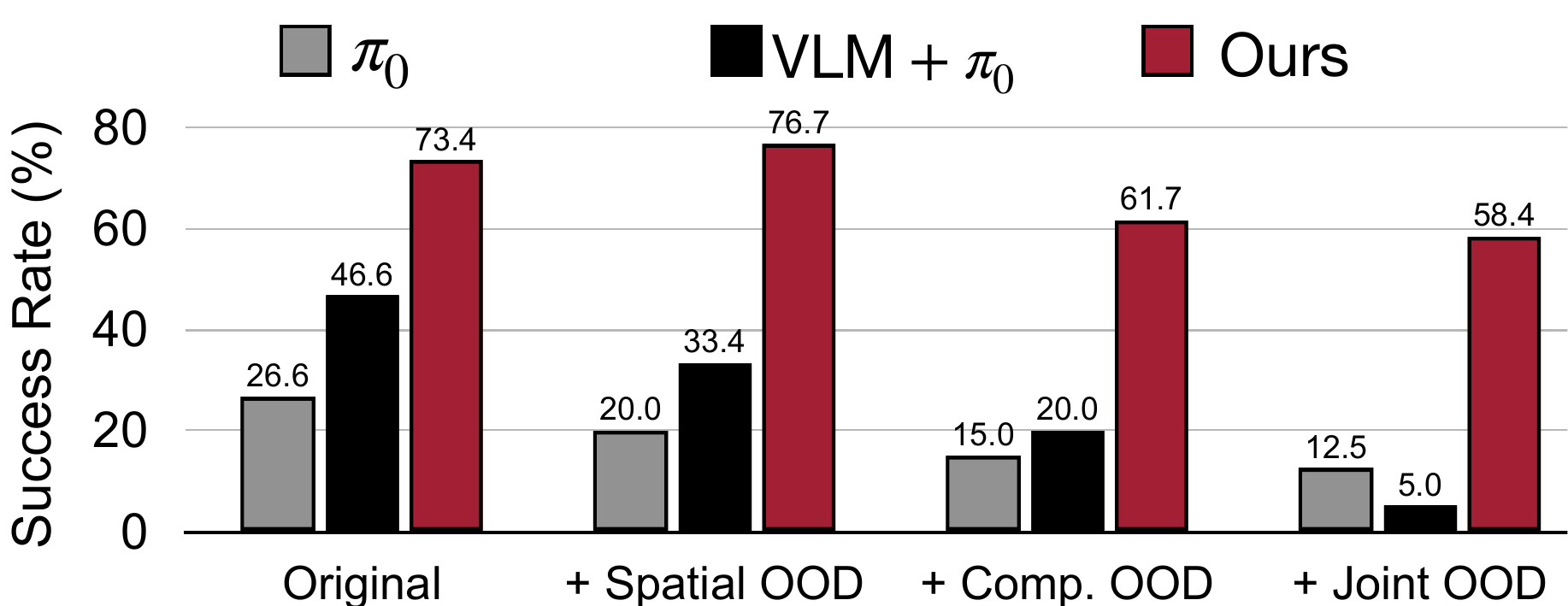}
  \caption{
    \textbf{Quantitative evaluation on OOD variants of the Pick\_Veg task.} The x-axis denotes increasingly difficult evaluation settings. Specifically, \texttt{Original} refers to the in-distribution setting. \texttt{+ Spatial OOD} introduces spatial permutations of objects that is unseen in the training data. \texttt{+ Comp.OOD} assesses compositional generalization by requiring the model to manipulate two objects, whereas training only involves single-object tasks. Finally, \texttt{+ Joint OOD} combines both spatial and compositional shifts.
}
  \label{fig:ood_bar_fig}
\end{figure}

We design OOD (Out-Of-Distribution) variants of the Pick\_Veg task to evaluate the generalization capability of our \methodshort framework. As shown in~\fig{ood_bar_fig}, our \methodshort framework remains robust in OOD settings, whereas baseline models suffer a substantial performance drop. 

\begin{figure*}[ht]
  \centering
  \includegraphics[width=\linewidth]{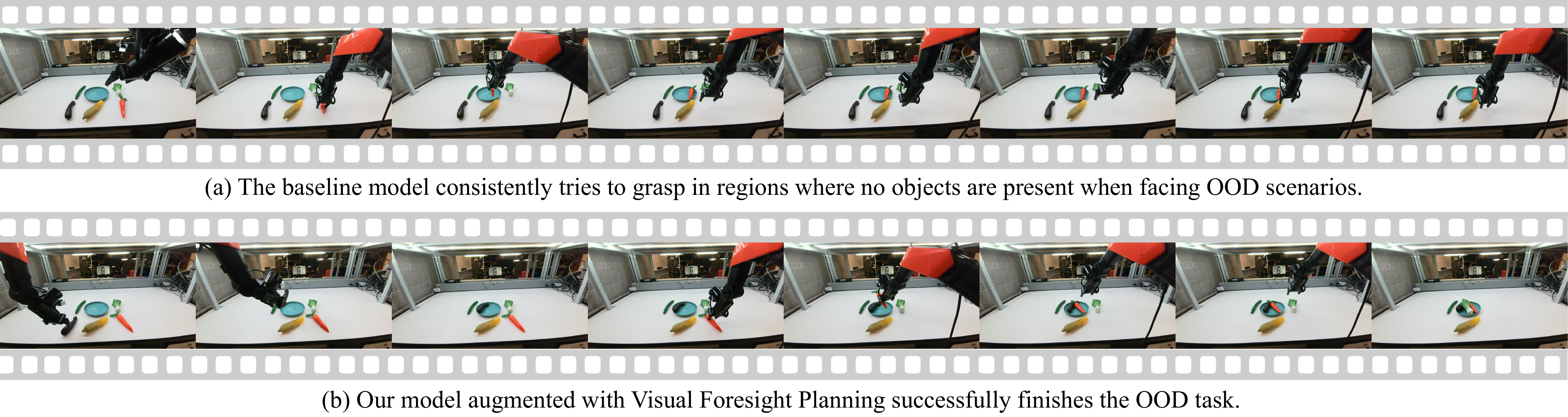}
  \caption{
    \textbf{Qualitative results for a compositional OOD task.}
    The overall instruction is to "put the carrot, bok choy, and eggplant into the plate," a sequence not seen in training.
    (a) The baseline (VLM + $\pi_0$) gets lost in this task, repeatedly trying to grasp the first object's location even after it is gone.
    (b) Our method robustly executes all three subtasks with the augmentation of visual guidance for each step.
}
  \label{fig:10_qualitative_results_ood}
  \vspace{-10pt}
\end{figure*}

To better illustrate this, we provide a qualitative example in~\fig{10_qualitative_results_ood}. The task requires the robot to put three different vegetables (a carrot, a bok choy, and an eggplant) into the plate, a composition of subtasks never seen during training, where only single-vegetable picking was demonstrated. As a result, the baseline model ($\text{VLM} + \pi_0$,~\fig{10_qualitative_results_ood} (a)) 
fails to adapt. It gets stuck in a loop, repeatedly attempting to grasp the same region where the carrot was, even though it is now empty. In contrast, the model augmented with our \methodshort framework successfully completes this complex OOD task, as shown in ~\fig{10_qualitative_results_ood} (b). Our framework correctly monitors the task progress and provides corresponding visual foresight, allowing the VLA to robustly execute all three distinct subtasks in sequence.

\subsection{Data Efficiency}
\label{supp:data_efficiency}

\begin{figure}[t]
    \centering
    \includegraphics[width=\linewidth]{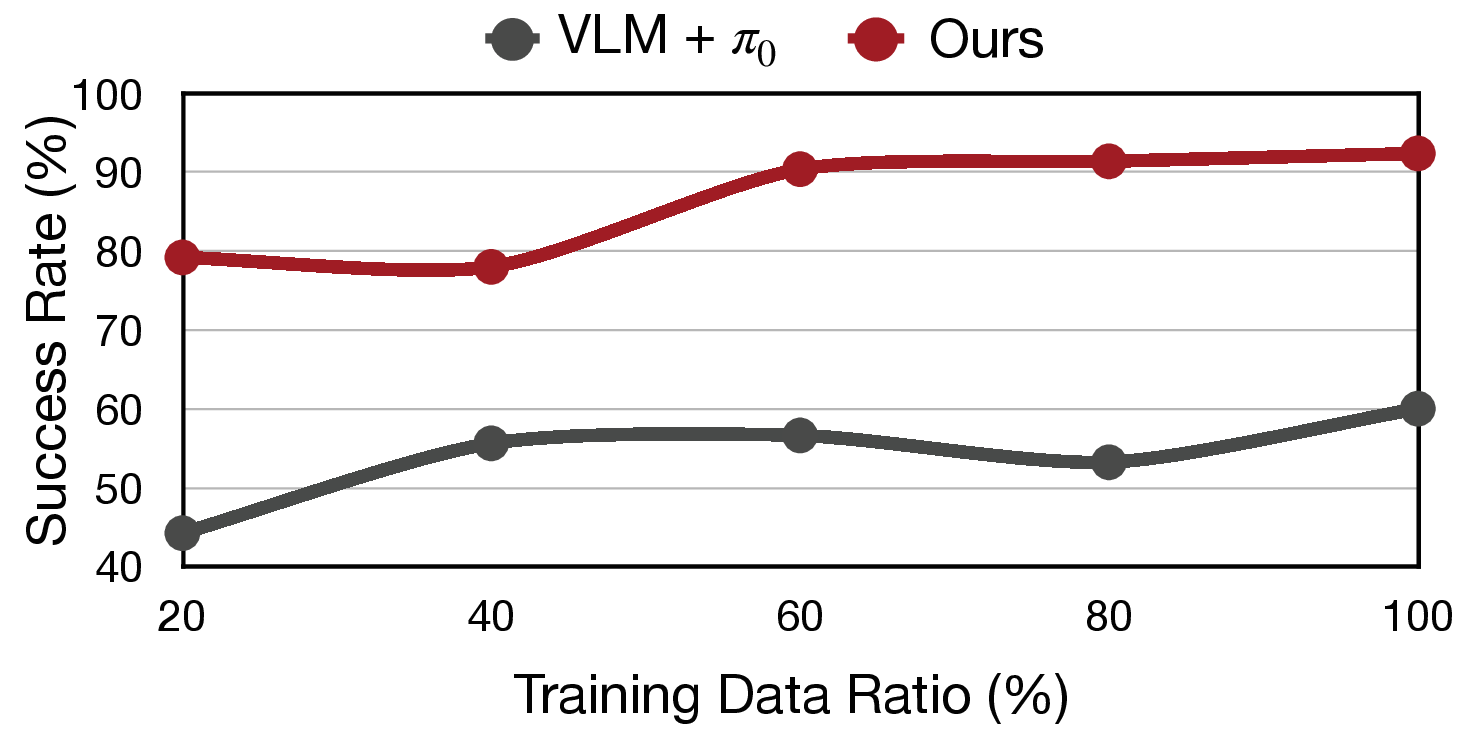}
    \vspace{-7pt}
    \caption{
        \textbf{Data efficiency analysis on the Clean\_Rubb task.} 
        We compare the success rates of our method (red) versus the VLM-augmented baseline (gray) when trained on data subsets ranging from 20\% to 100\%. Our method consistently outperforms the baseline across all data regimes.
    }
    \label{fig:data_efficiency}
  \end{figure}

We evaluate data efficiency on the challenging Clean\_Rubb task by varying the size of the training set. As shown in~\fig{data_efficiency}, our method exhibits strong sample efficiency: with only 60\% of the data, it attains a >90\% success rate, and with just 20\% of the data, it still achieves 79\%, substantially outperforming the baseline. Although the baseline method shows performance gains as the dataset grows, its rate of improvement is considerably slower, underscoring the superior data efficiency of our \methodshort approach. Note that for this ablation, we construct training subsets solely from the Clean\_Rubb task data. Therefore, the success rate at 100\% data usage here is slightly lower than the main benchmark result reported in the main paper, which benefits from multi-task co-training.


Furthermore, limited fine-tuning data severely constrains baseline performance. As shown in Fig.~\ref{fig:8_main_results}, while our system achieves a 93.4\% success rate in the Pick\_Tool task, baselines fall below 30\%. Since there are 5 distinct tools forming $5! = 120$ spatial permutations, and the task requires manipulating any one of these five tools within each layout, the total number of unique configurations reaches $120 \times 5 = 600$. Collecting teleoperation trajectories for every permutation within the finetuning data for the single task is sub-optimal and practically infeasible. Therefore, we hypothesize that baselines fail to generalize in this data-scarce regime because they struggle to ground task prompts to specific tool objects. 


To further investigate this, Table~\ref{tab:tool_pick_with_loc} presents an ablation where semantic instructions (e.g., "Pick up the hammer") are replaced with spatial ones (e.g., "Pick up the object on the bottom right") in both finetuning and inference. This modification significantly improves baseline accuracy by decoupling semantic recognition from action execution. By bypassing the need for the VLA to infer object semantics from visual observations, the model can dedicate its limited capacity to action prediction. Nevertheless, these baselines still lag behind our \methodshort framework, as spatial text remains a coarser and more ambiguous modality compared to the fine-grained image-based guidance we introduce. These results collectively suggest that our visual foresight mechanism fundamentally enhances data efficiency: by converting abstract semantic instructions into concrete visual targets, it reduces the reliance on exhaustive demonstration coverage and enables robust generalization even when fine-tuning data is far too sparse to cover the full combinatorial space of task configurations.

\begin{table}[t]
\centering
\caption{\textbf{Ablation on instruction modalities for the Pick\_Tool task}. We compare standard semantic instructions with spatial-based text prompts. "Spatial Text" provides location of the target object, simplifying the task by removing the need for object grounding.}
\label{tab:tool_pick_with_loc}
\scalebox{0.88}{
\begin{tabular}{ccc}
\toprule
Method & Instruction Type & Success Rate (\%) $\uparrow$ \\ \midrule
Single $\pi_0$  & Semantic Text & 20.0 \\
Single $\pi_0$  & Spatial Text  & 46.8 \\ 
Ours & Semantic Text + Goal Image & 93.4 \\ \bottomrule
\end{tabular}
}
\end{table}

\section{Conclusion}
\label{sec:conclusion}

In this work, we present \method (\methodshort), a general and efficient planner that instructs VLA models with step-by-step imagined future observations and subtask guidance. By providing imagined future observations, it allows VLAs to focus on visuo-motor execution, leading to improved accuracy and generalization. The foresight image generation inside our planner is efficient, running at 0.33s to produce a 640×480 future observation on an H100 GPU. We evaluate our method across 11 diverse real-world tasks, it achieves an average success rate of 87.4\%, outperforming both vanilla and VLM-augmented baselines by large margins. Furthermore, it demonstrates strong performance in out-of-distribution scenarios. Overall, our results show that \methodshort is a powerful and practical direction for enabling more capable and reliable embodied intelligence.

\section*{Acknowledgements}
We thank MIT-IBM Watson AI Lab, Amazon and National Science Foundation for supporting this research. We
thank NVIDIA for donating the DGX server.

{
    \small
    \bibliographystyle{ieeenat_fullname}
    \bibliography{main}
}

\newpage
\appendix

\section{Appendix}

\subsection{Table of Contents}
\begin{itemize}
    \item Sec.~\ref{supp:qualitative_foresight}: We present additional qualitative foresight generation results, particularly in out-of-distribution scenarios to highlight the strong generalization of our model.
    \item Sec.~\ref{supp:task_desp}: The detailed descriptions of all our real-world tasks with visual examples.
    \item Sec.~\ref{supp:vlm_eval}: Evaluation on different VLMs' performance as the high-level task planner.
    \item Sec.~\ref{supp:vlm_prompt}: The prompt we use for the VLM model.
    \item Sec.~\ref{supp:demo_video}: A video demo.
\end{itemize}

\subsection{Qualitative Foresight Generation Results}
\label{supp:qualitative_foresight}
\begin{figure*}[h]
    \centering
    \includegraphics[width=\linewidth]{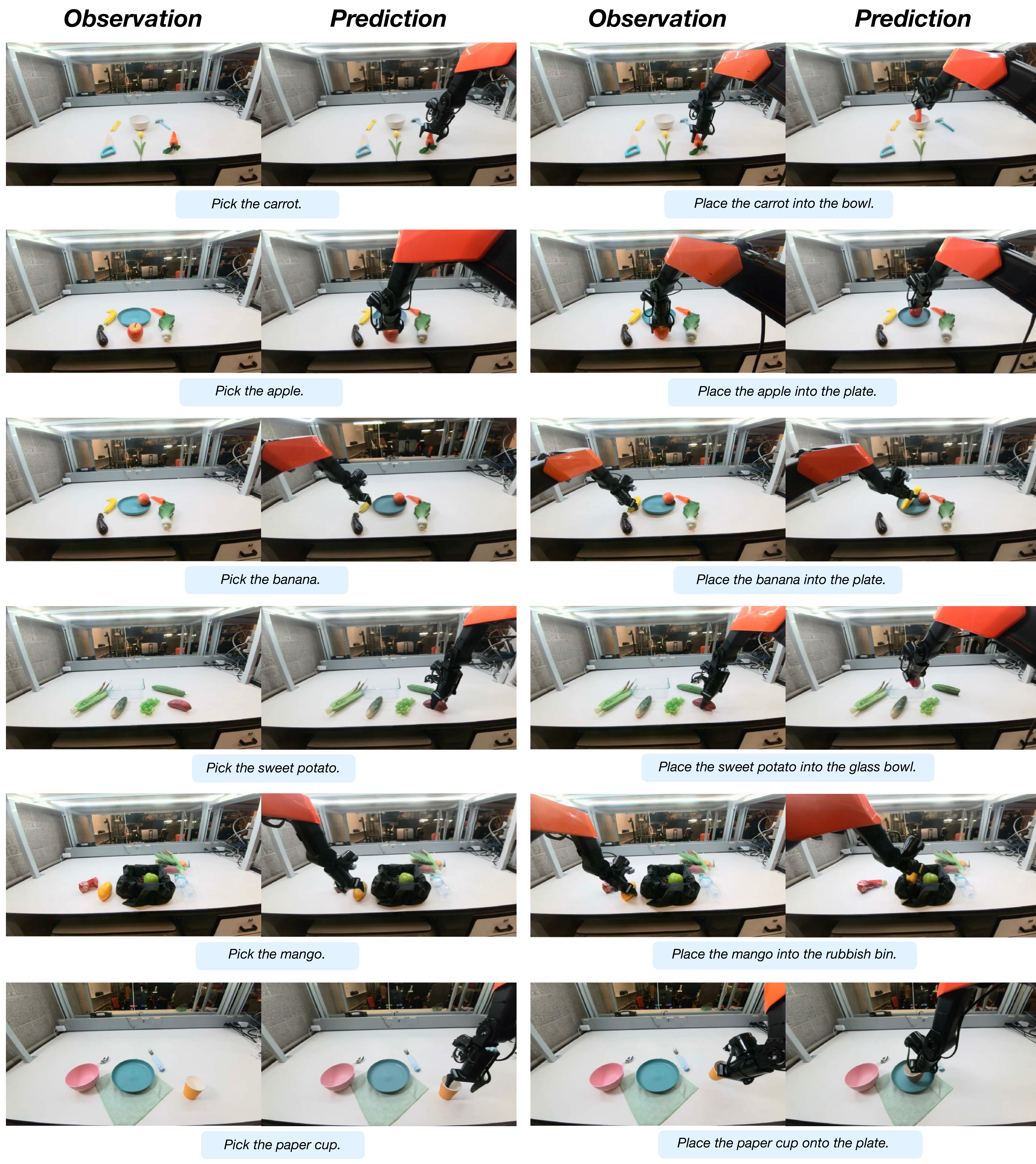}
    \caption{
        \textbf{Qualitative Foresight Image Generation Results.}
    }
    \label{fig:ood_supp}
\end{figure*}

Figure~\ref{fig:ood_supp} presents further qualitative results of our foresight image generation model. All scenarios shown are out-of-distribution, highlighting the strong capability and generalization of our model. This further demonstrates the effectiveness of our large-scale pretraining. We provide a detailed discussion of these scenarios below.

\begin{itemize}
    \item \textbf{First Row}: While all the objects are present in our dataset, they have never been co-located within the same scene. Although the dataset contains bowls, we do not include any scenes in which objects are placed inside a bowl.
    \item \textbf{Second Row}: This scenario is similar to Pick\_Veg; however, we introduce fruits (an apple and a banana) into the scene. Notably, neither apples nor bananas are present in our dataset.
    \item \textbf{Third Row}: There is already an apple on the plate. In the Pick\_Veg task, however, we place only a single object onto the plate.
    \item \textbf{Fourth Row}: This scenario represents a completely unseen setting: all objects, including the new vegetables, fruits, and the glass bowl, are not present in our dataset.
    \item \textbf{Fifth Row}: This scenario is similar to Clean\_Rubb; however, instead of placing rubbish into the bin, we place fruits into it.
    \item \textbf{Sixth Row}: This task is entirely novel and includes new objects such as a knife, a fork, and a paper cup. The paper cup has never been positioned on a plate in our dataset.
\end{itemize}



\subsection{Detailed Task Descriptions}
\label{supp:task_desp}

In Table~\ref{tab:task_descriptions} and ~\ref{tab:task_descriptions_2}, we provide a comprehensive breakdown, with visual examples, of the 11 real-world robotic tasks used in our evaluation suite. To rigorously evaluate the models' performance, we intentionally introduce distributional shifts between training and inference in some of the testbenches, such as increasing the number of objects, randomizing initial poses, or introducing novel distractors. Please refer to the tables for a summary of the operational protocol and data distribution for each task.

\begin{figure}[t]
    \centering
    \includegraphics[width=\linewidth]{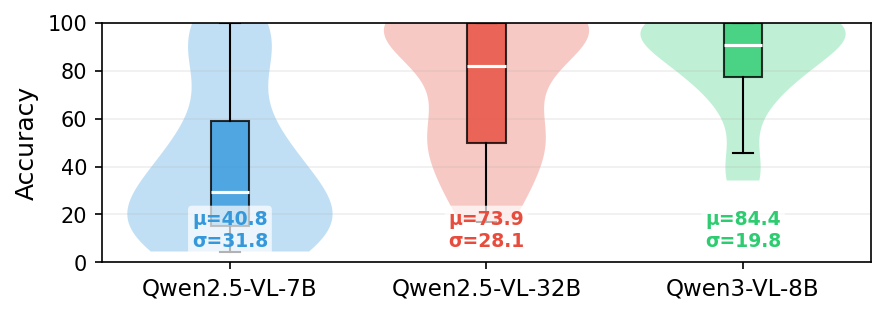}
    \caption{\textbf{Subtask Planning Accuracy across Different VLMs.}}
    \label{fig:rebut:vlm_eval}
    \vspace{-10pt}
\end{figure}

\begin{table}[t]
    \centering
    \caption{\textbf{The prompt template used for the VLM planner.} The variable \texttt{\{Task\}} represents the high-level human instruction.}
    \small
    \fbox{%
    \begin{minipage}{0.95\linewidth}
        \vspace{0.2cm}
        \textbf{SYSTEM / INITIAL PROMPT ($t=0$)}
        \vspace{0.1cm}
        \hrule
        \vspace{0.2cm}
        
        You are a robot controller. Please plan to finish the task in several steps. And give instruction for each step in a concise way. \\
        The task is to ``\texttt{\{Task\}}''.
        
        \vspace{0.2cm}
        \textbf{RULES:}
        \begin{itemize}[leftmargin=*, nosep]
            \item During the job, I will continuously give you an observation image of the current state.
            \item Based on the observation, please judge if the last instruction has been finished.
            \begin{itemize}[nosep]
                \item If yes, give me the instruction for the next step.
                \item If no, repeat the instruction of the ongoing subtask.
            \end{itemize}
            \item You're not required to describe the observation. Only output the instruction for each subtask.
        \end{itemize}
        
        \vspace{0.2cm}
        Now, you are only required to output instruction for the first step.
        
        \vspace{0.2cm}
        \textbf{VISUAL INPUT}: \texttt{[Initial Observation Image]}
        \vspace{0.2cm}
    \end{minipage}%
    }
    
    
    \fbox{%
    \begin{minipage}{0.95\linewidth}
        \vspace{0.2cm}
        \textbf{FOLLOW-UP PROMPT ($t > 0$)}
        \vspace{0.1cm}
        \hrule
        \vspace{0.2cm}
        
        \textbf{VISUAL INPUT}: \texttt{[Current Observation Image]}
        \vspace{0.2cm}
        
        Pay attention to the latest observation. Firstly, judge if the last instruction has been finished. Secondly, if yes, give me the instruction for the next step; if no, repeat the instruction of the ongoing subtask. \\
        Your answer should be concise and deterministic. \\
        Remember, your Overall Task is ``\texttt{\{Task\}}''.
        \vspace{0.2cm}
    \end{minipage}%
    }

    \label{tab:prompts}
\end{table}

\newcolumntype{Y}{>{\raggedright\arraybackslash}X}

\begin{table*}[h]
\centering
\caption{\textbf{Overview of robot tasks and descriptions in our real-world dataset} (Part I).}
\label{tab:task_descriptions}
\renewcommand{\arraystretch}{2.5}
\begin{tabularx}{\textwidth}{@{} m{0.3\textwidth} Y @{}} 
\toprule
\textbf{Task Example} & \textbf{Detailed Description} \\ \midrule

\begin{minipage}[t]{0.3\textwidth}
    \centering
    \vspace{-15pt}
    \includegraphics[width=0.8\linewidth]{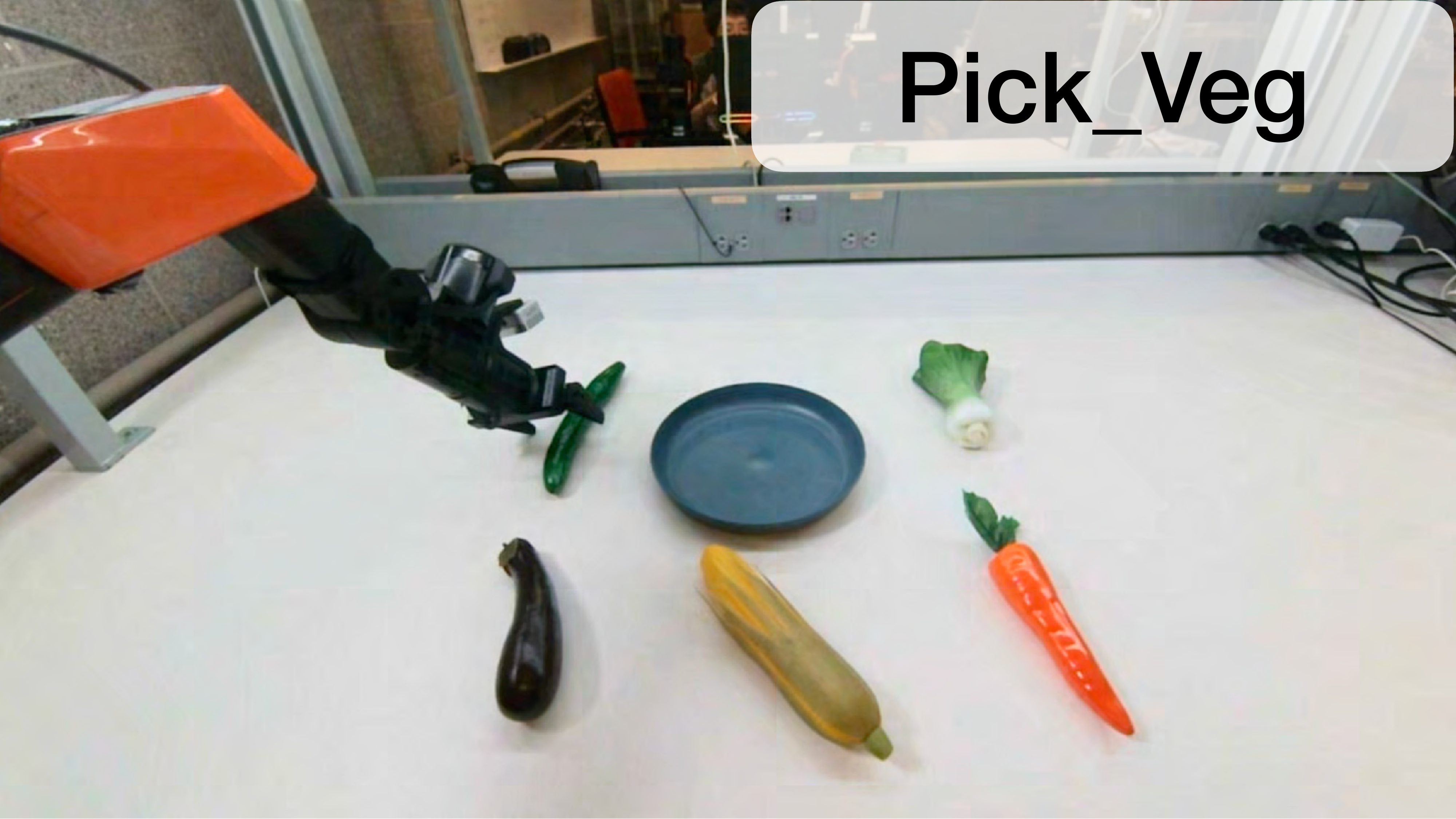} \hspace{15pt} \\
    \vspace{5pt}
\end{minipage} & 
\textbf{Protocol:} The robot must identify and pick a specific vegetable from a set of five and place it onto a plate. 
\smallskip \newline \textbf{Data:} Fine-tuning episodes exclusively feature single-object manipulation. 
Evaluation follows this setting unless otherwise specified (e.g., Fig.~\ref{fig:ood_bar_fig}).
\\ \midrule

\begin{minipage}[t]{0.3\textwidth}
    \centering
    \vspace{-15pt}
    \includegraphics[width=0.8\linewidth]{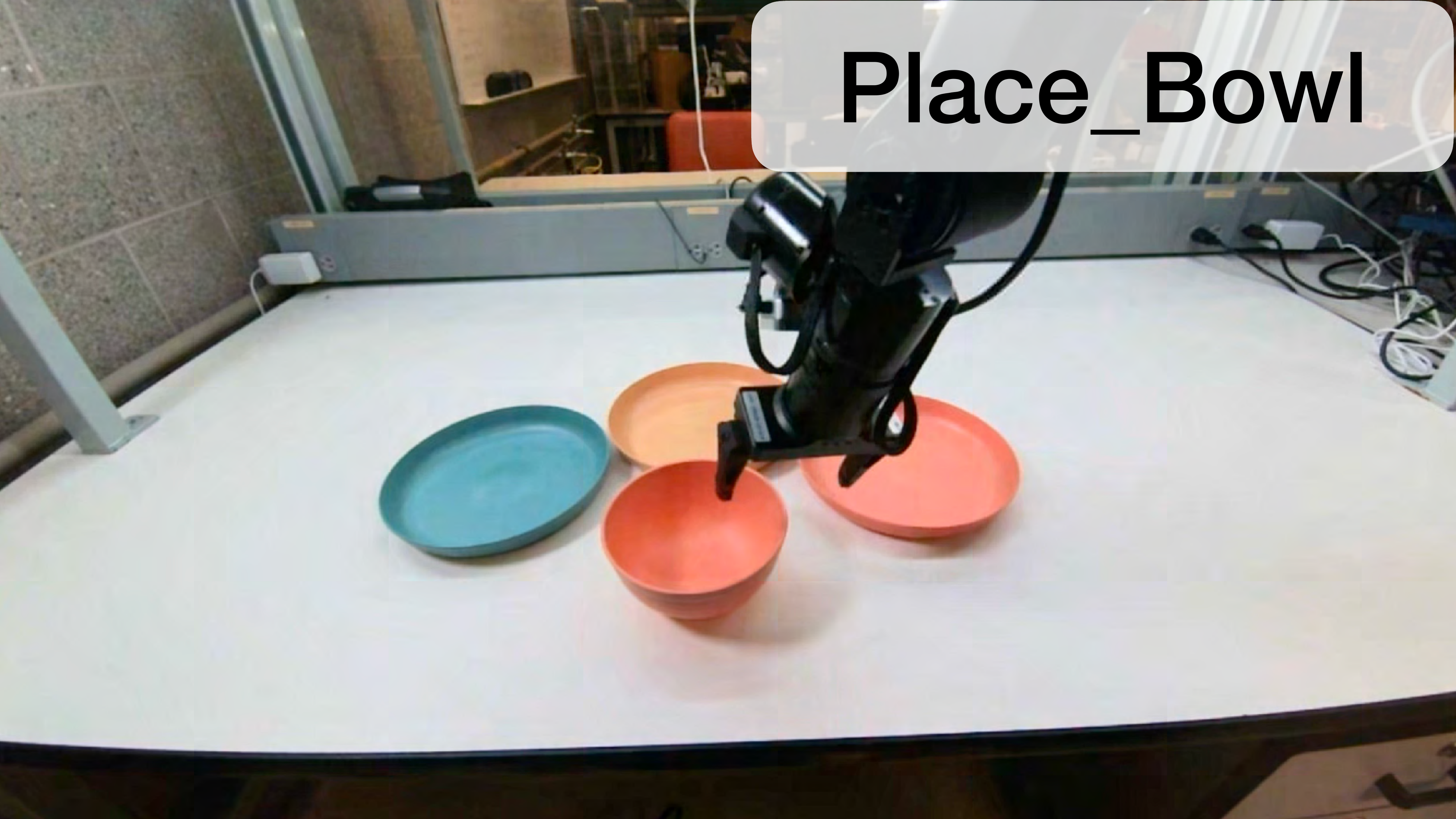} \hspace{15pt} \\
    \vspace{5pt}
\end{minipage} & 
\textbf{Protocol:} The robot is required to pick a bowl and place it onto a matching color-coded plate. 
\smallskip \newline \textbf{Data:} Both training and inference setups involve a single bowl and three plates on the workspace. The presence of a color-matched plate is guaranteed in every episode.
\\ \midrule

\begin{minipage}[t]{0.3\textwidth}
    \centering
    \vspace{-15pt}
    \includegraphics[width=0.8\linewidth]{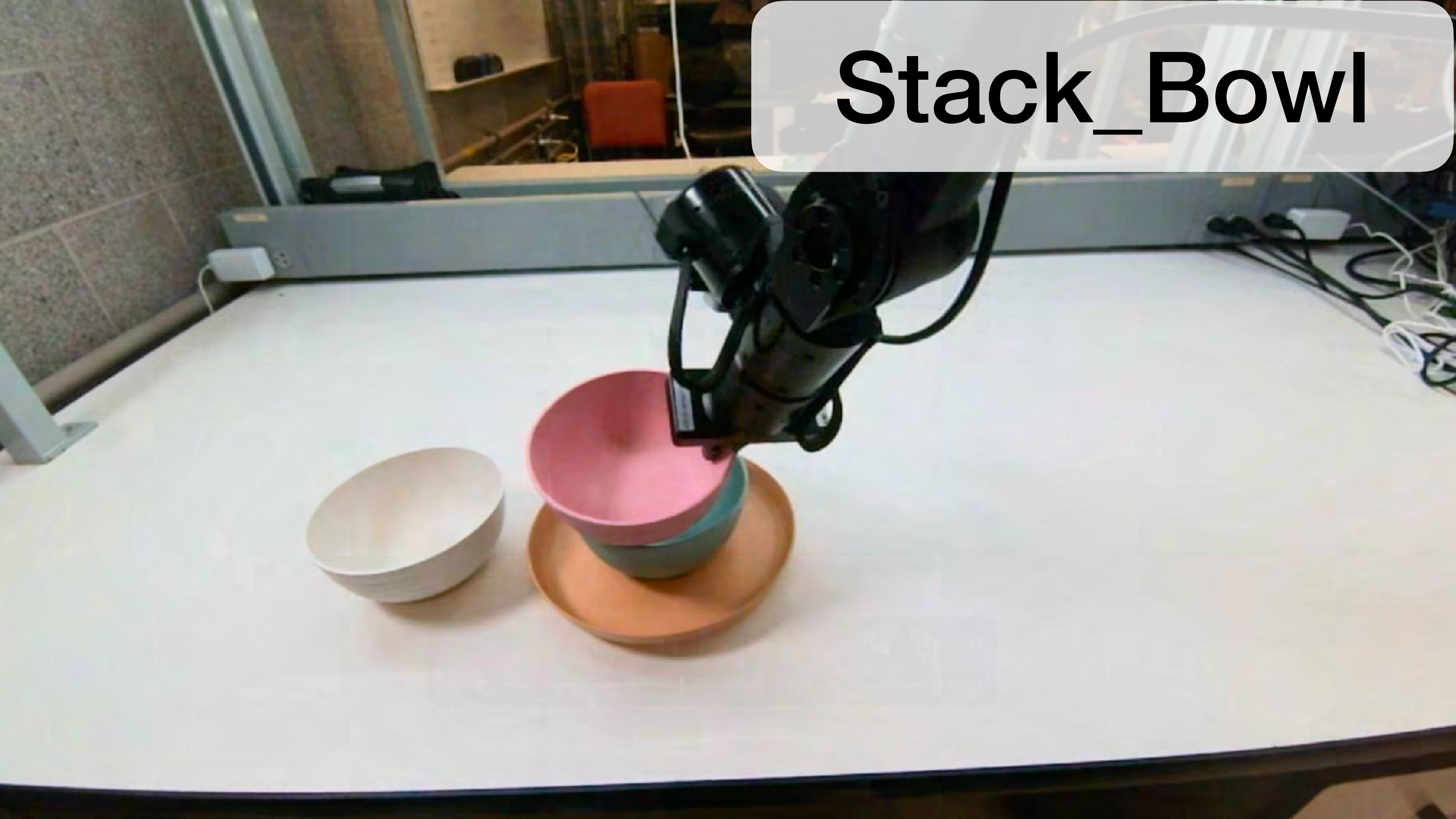} \hspace{15pt} \\
    \vspace{5pt}
\end{minipage} & 
\textbf{Protocol:} The robot is required to pick up bowls one by one and stack them onto the plate. 
\smallskip \newline \textbf{Data:} Both training and inference involve stacking three bowls. During evaluation, the bowls can either be pre-placed on the desk or introduced at any time.
\\ \midrule

\begin{minipage}[t]{0.3\textwidth}
    \centering
    \vspace{-15pt}
    \includegraphics[width=0.8\linewidth]{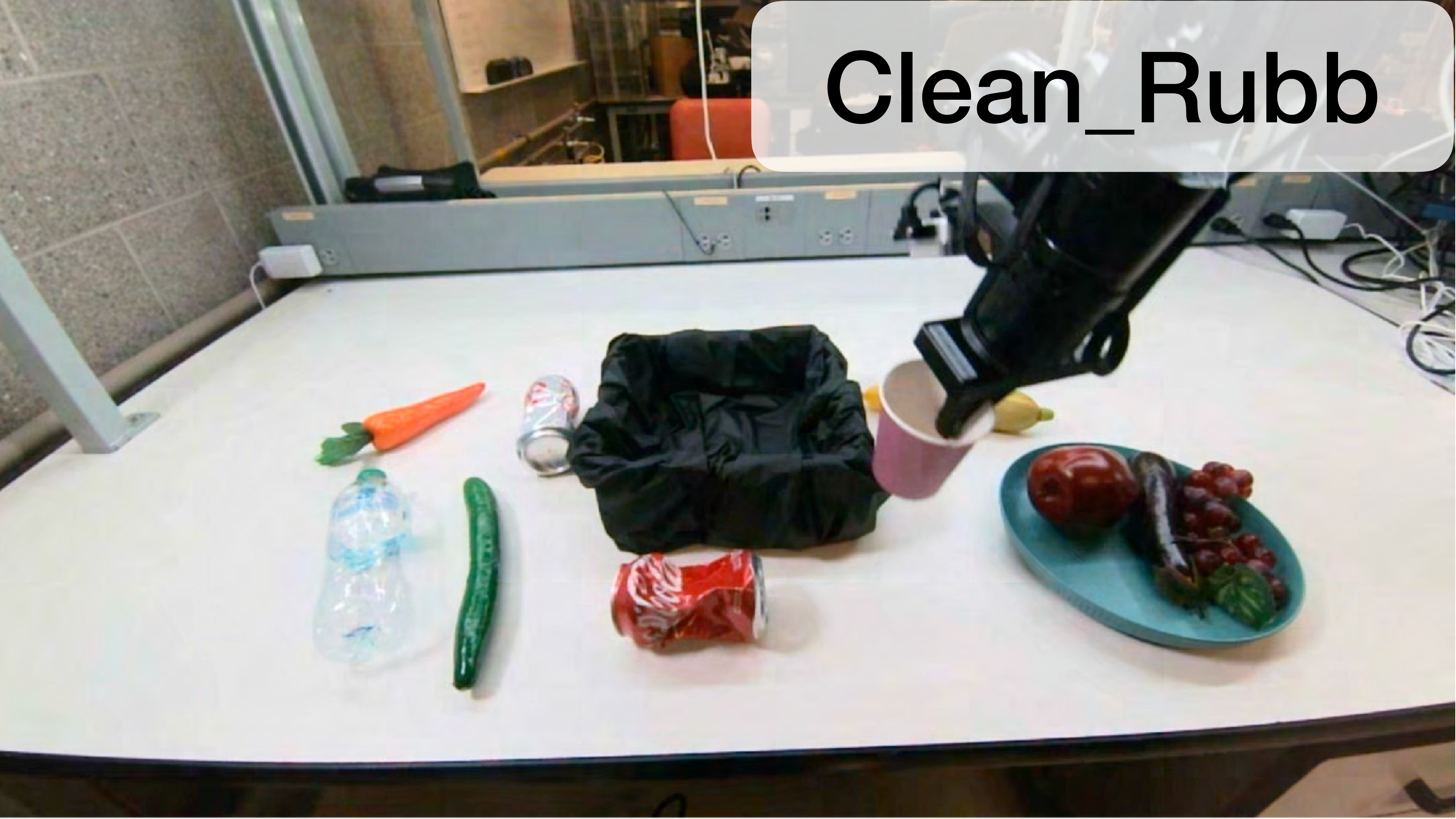} \hspace{15pt} \\
    \vspace{5pt}
\end{minipage} & 
\textbf{Protocol:} The robot is required to pick up rubbish items and dispose of them into a bin. 
\smallskip \newline \textbf{Data:} Items are randomly positioned on the workspace during both training and inference. Background settings are varied across episodes.
\\ \midrule

\begin{minipage}[t]{0.3\textwidth}
    \centering
    \vspace{-15pt}
    \includegraphics[width=0.8\linewidth]{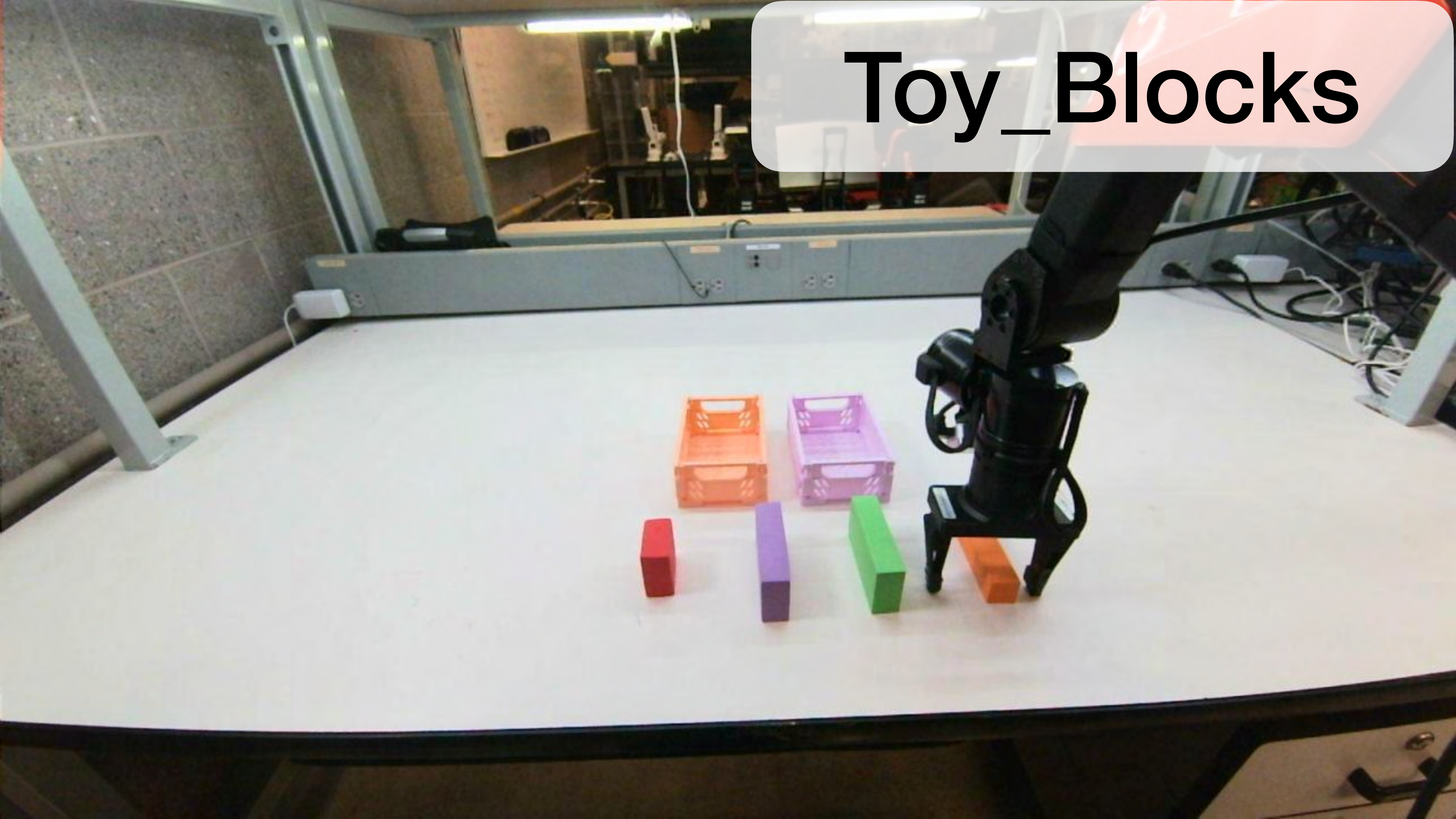} \hspace{15pt} \\
    \vspace{5pt}
\end{minipage} & 
\textbf{Protocol:} The robot sorts various toy blocks into boxes with matching colors. 
\smallskip \newline \textbf{Data:} The workspace contains 2 boxes and 4 blocks of different shapes and colors. While training episodes involve picking two blocks, evaluation requires sorting 2 to 3 blocks.
\\ \midrule

\begin{minipage}[t]{0.3\textwidth}
    \centering
    \vspace{-15pt}
    \includegraphics[width=0.8\linewidth]{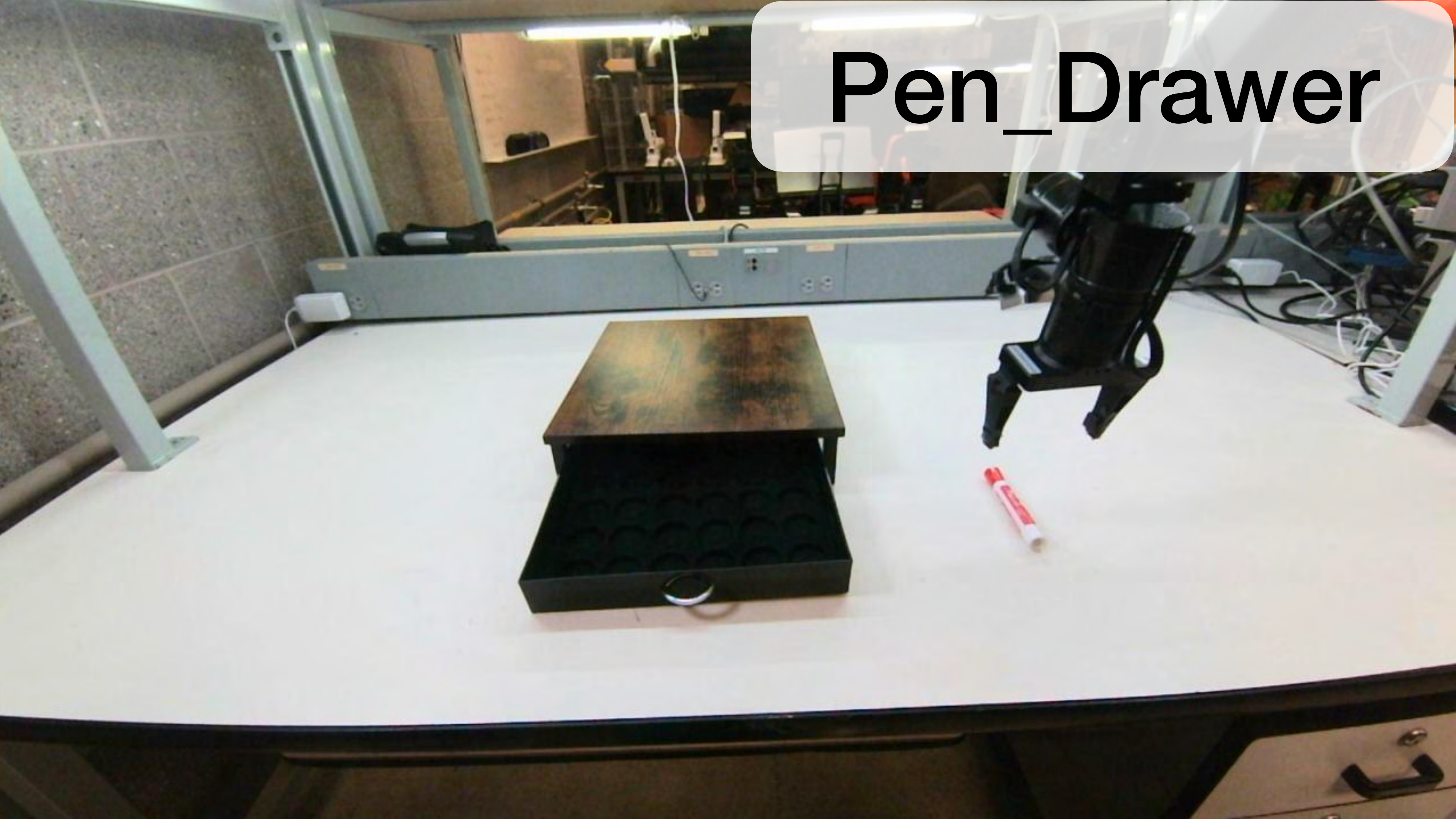} \hspace{15pt} \\
    \vspace{5pt}
\end{minipage} & 
\textbf{Protocol:} The robot picks up pens from the desk, places them inside a drawer, and then closes the drawer. 
\smallskip \newline \textbf{Data:} Training data exclusively involves manipulating a single pen. Evaluation requires the robot to place two pens into the drawer.
\\ \midrule

\begin{minipage}[t]{0.3\textwidth}
    \centering
    \vspace{-15pt}
    \includegraphics[width=0.8\linewidth]{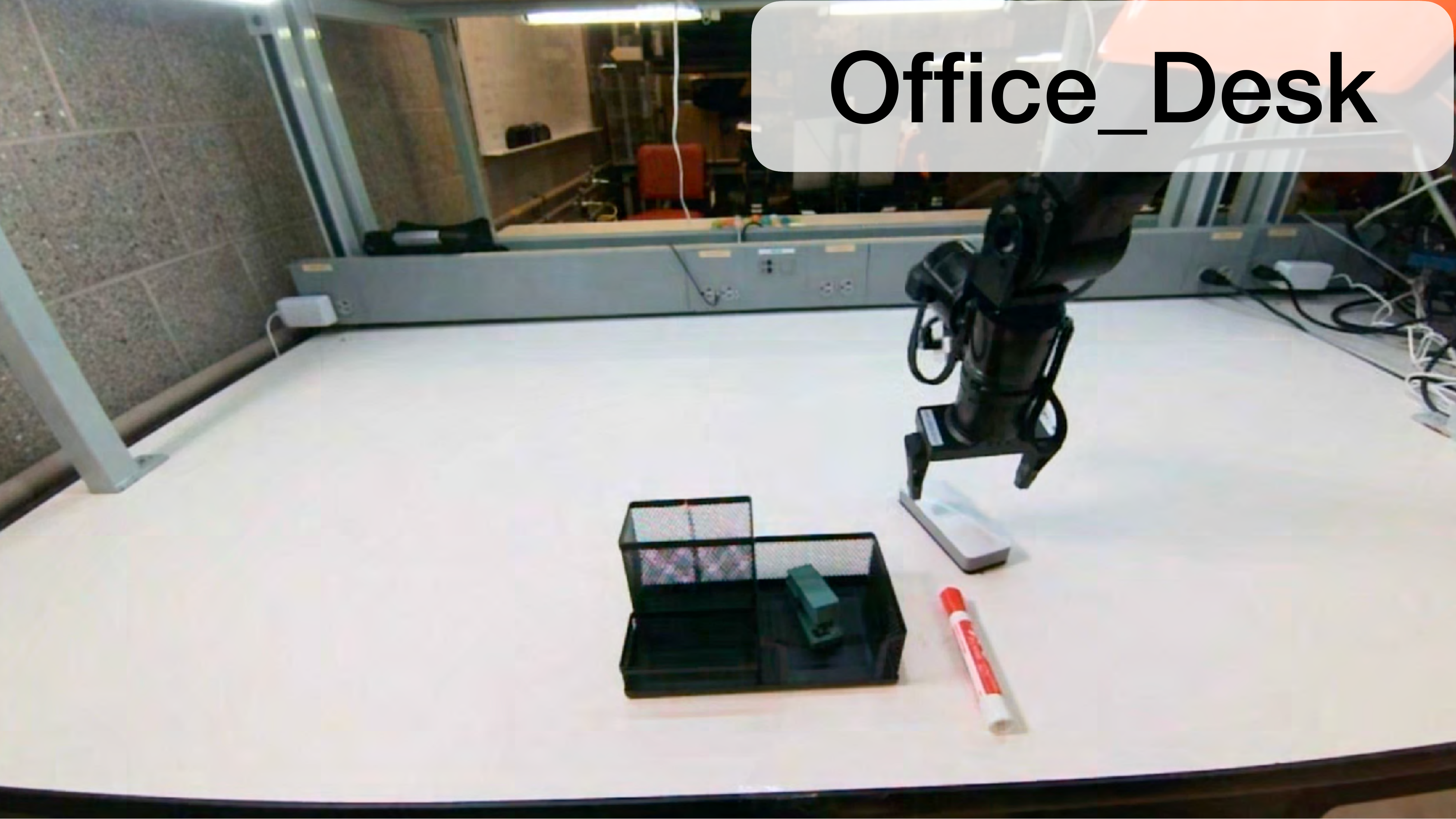} \hspace{15pt} \\
    \vspace{5pt}
\end{minipage} & 
\textbf{Protocol:} The robot is required to organize a cluttered desk by placing items into a designated organizer. 
\smallskip \newline \textbf{Data:} The task involves individual object placements, including a whiteboard eraser, a stapler, and a pen. The robot is required to sort all three objects in both training and evaluation.
\\ \midrule
\bottomrule
\end{tabularx}
\end{table*}

\begin{table*}[t]
\centering
\caption{\textbf{Overview of robot tasks and descriptions in our real-world dataset} (Part II).}
\label{tab:task_descriptions_2}
\renewcommand{\arraystretch}{2.5}
\begin{tabularx}{\textwidth}{@{} m{0.3\textwidth} Y @{}} 
\toprule
\textbf{Task Example} & \textbf{Detailed Description} \\ \midrule

\begin{minipage}[t]{0.3\textwidth}
    \centering
    \vspace{-15pt}
    \includegraphics[width=0.8\linewidth]{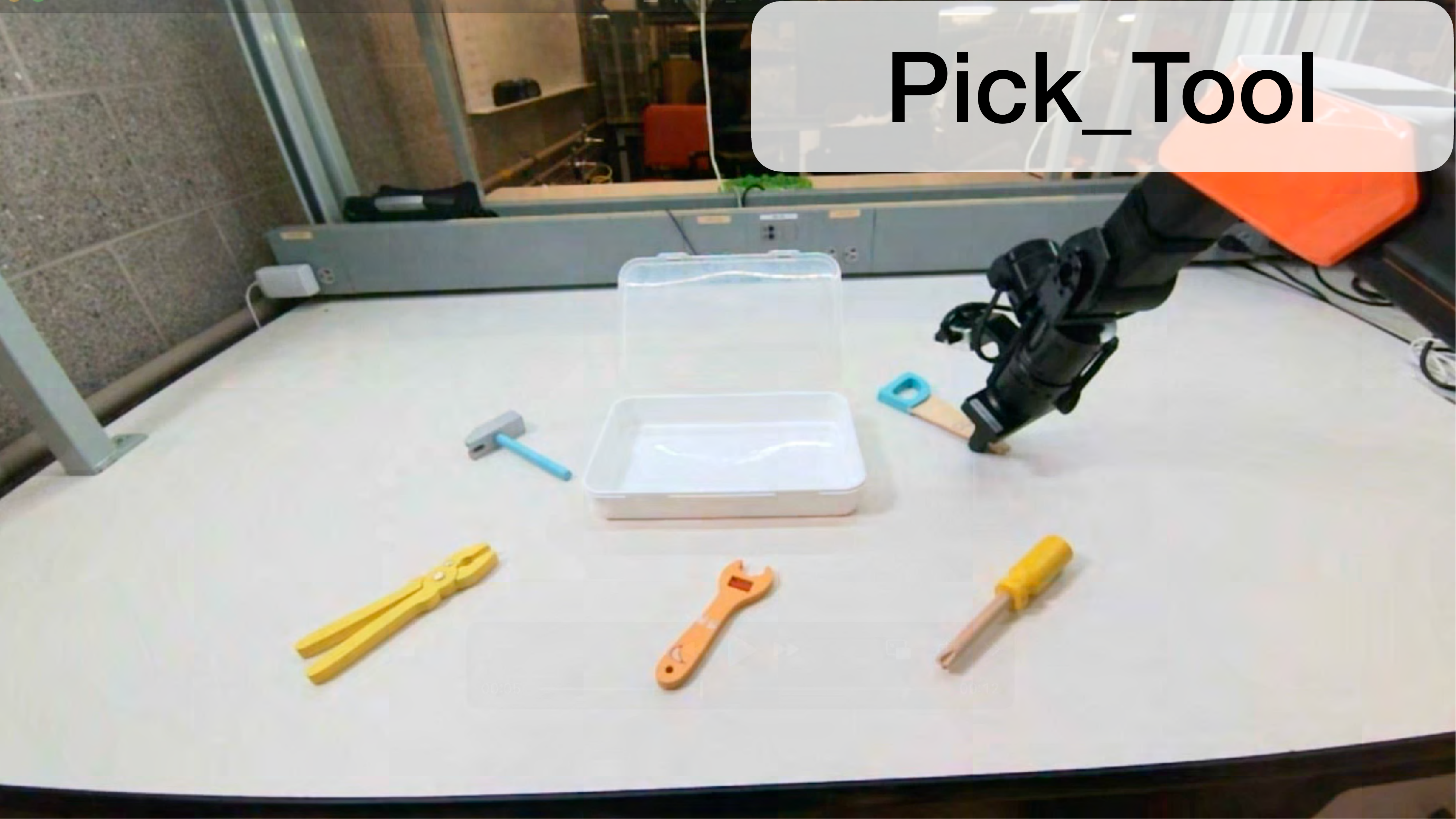} \hspace{15pt} \\
    \vspace{5pt}
\end{minipage} & 
\textbf{Protocol:} The robot identifies a specific tool and places it into a target container. 
\smallskip \newline \textbf{Data:} Both training and evaluation involve single-object manipulation. The task includes 5 distinct tools (hammer, hand saw, screwdriver, pliers, and wrench).
\\ \midrule

\begin{minipage}[t]{0.3\textwidth}
    \centering
    \vspace{-15pt}
    \includegraphics[width=0.8\linewidth]{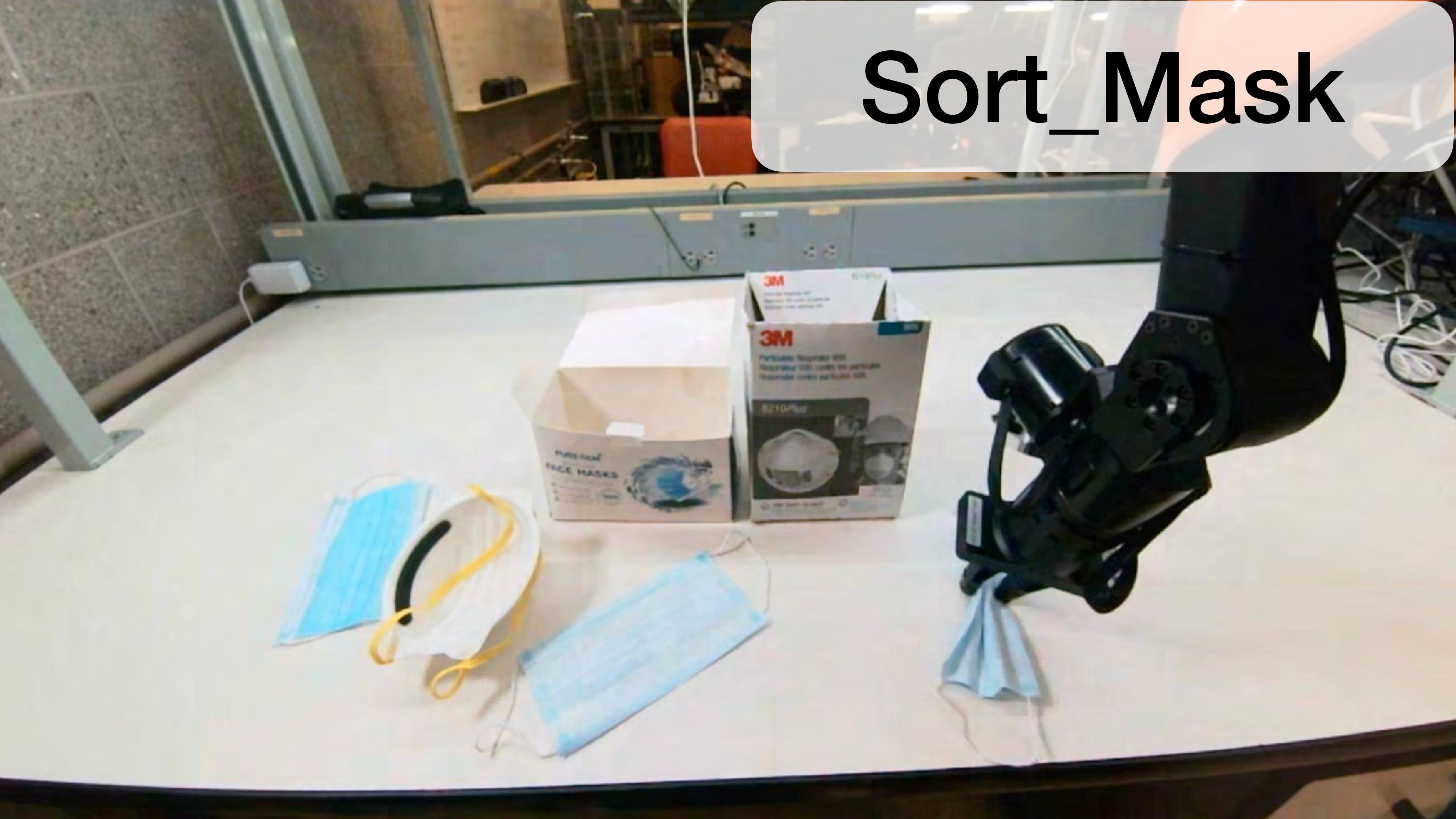} \hspace{15pt} \\
    \vspace{5pt}
\end{minipage} & 
\textbf{Protocol: }The robot distinguishes between N95 and surgical masks, sorting them into separate boxes.
\smallskip \newline \textbf{Data:} Both training and evaluation involve two types of masks. The positions of masks are randomly initialized.
\\ \midrule

\begin{minipage}[t]{0.3\textwidth}
    \centering
    \vspace{-15pt}
    \includegraphics[width=0.8\linewidth]{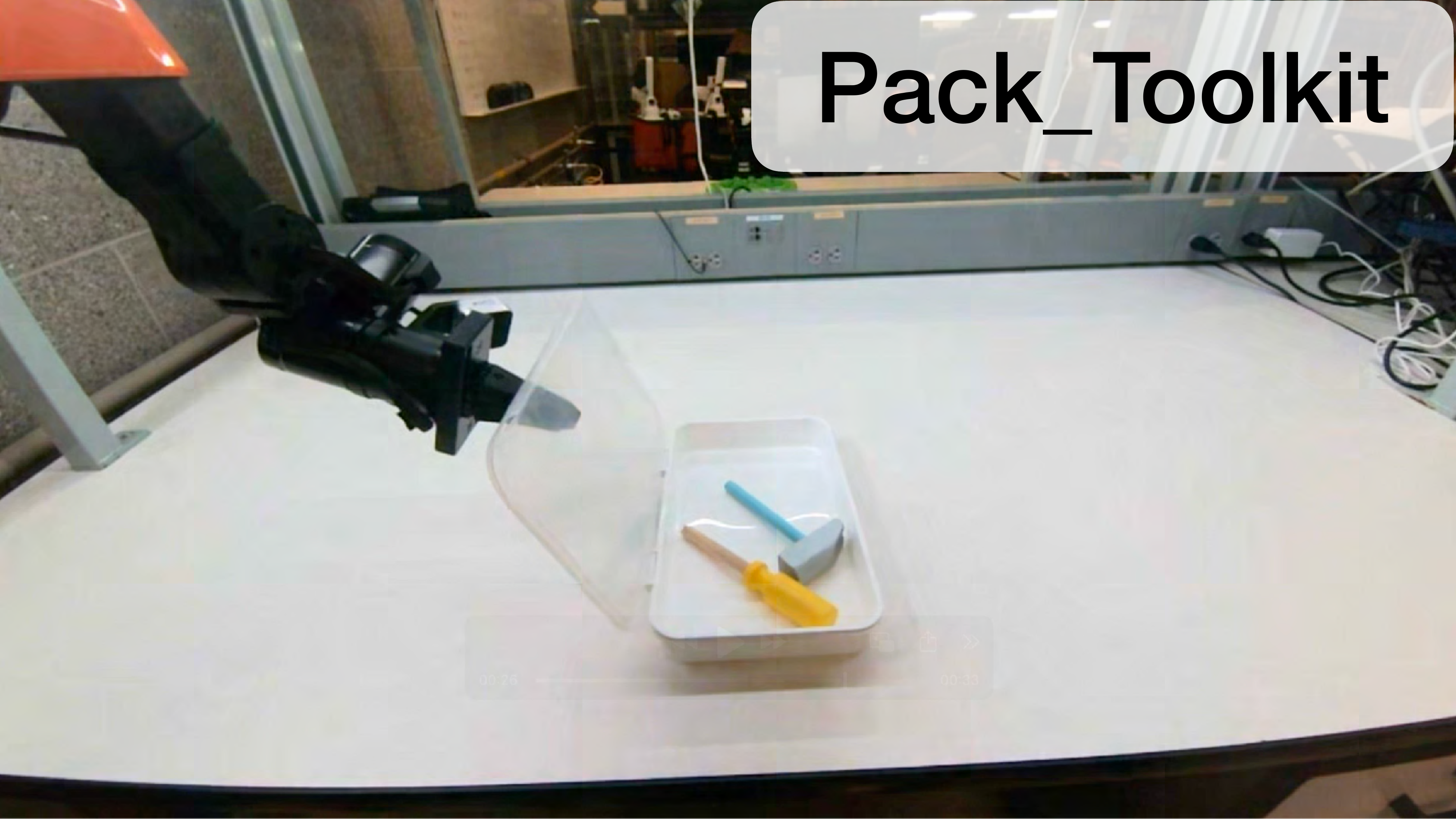} \hspace{15pt} \\
    \vspace{5pt}
\end{minipage} & 
\textbf{Protocol: }The robot retrieves designated tools, put them in the toolkit, and then closes the toolkit lid.
\smallskip \newline \textbf{Data:} Both training and evaluation involve two target objects. Unseen objects were introduced during the evaluation to test zero-shot abilities.
\\ \midrule

\begin{minipage}[t]{0.3\textwidth}
    \centering
    \vspace{-15pt}
    \includegraphics[width=0.8\linewidth]{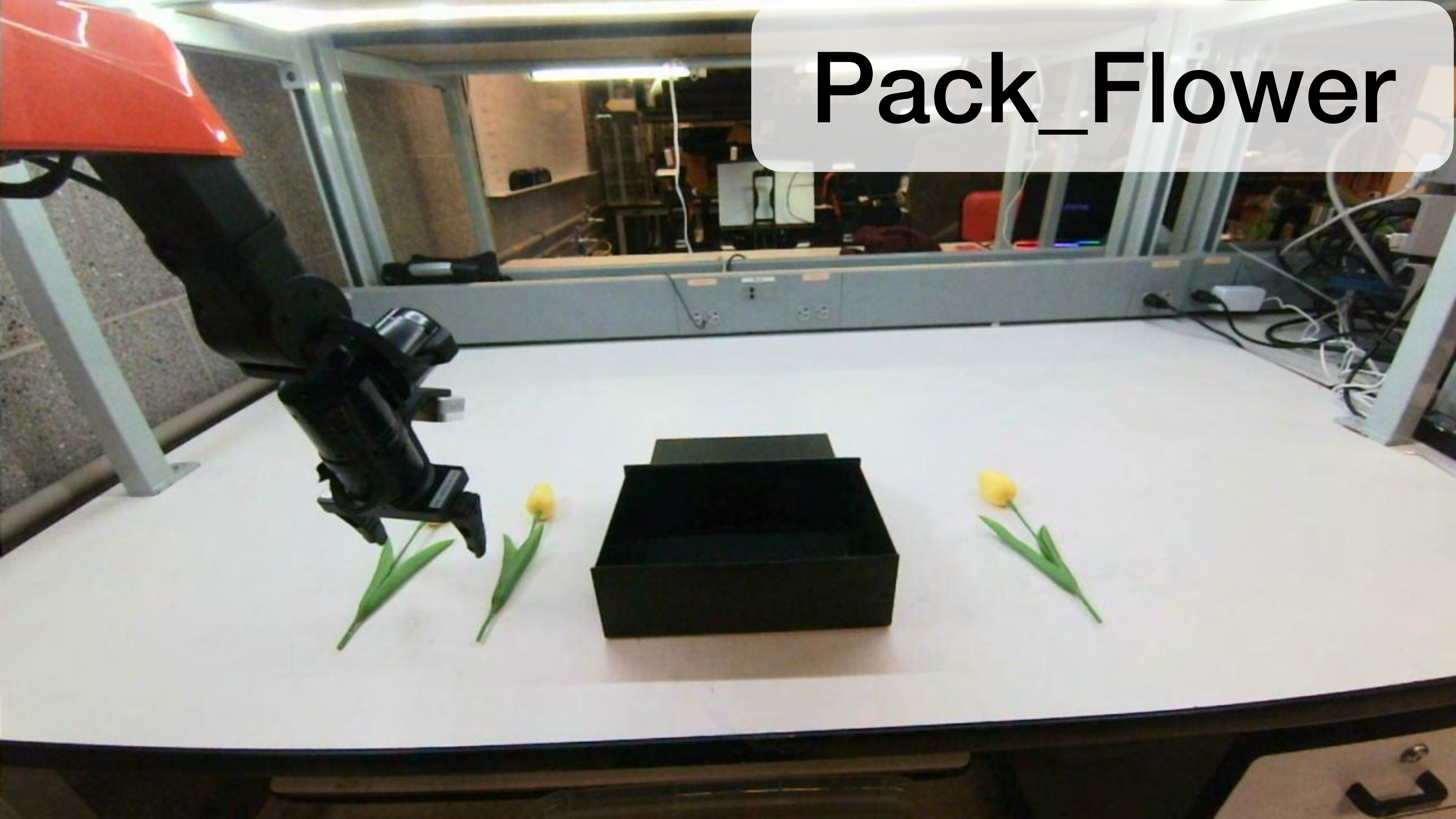} \hspace{15pt} \\
    \vspace{5pt}
\end{minipage} & 
\textbf{Protocol: }The robot picks up flowers on the desk and place them into a box.
\smallskip \newline \textbf{Data:} Both training and evaluation involve 2 to 3 flowers. Additional distractors, such as corn and grapes, are added to the workspace during evaluation.
\\ \midrule

\bottomrule
\end{tabularx}
\end{table*}

\subsection{Evaluating Performance of VLMs}
\label{supp:vlm_eval}


To assess the generalization capability of our planner, we employ a rigorous LLM-as-a-Judge protocol. Specifically, we utilize Gemini-3-pro-preview~\cite{Gemini3} to evaluate the semantic correctness of the VLM-predicted subtasks on our long-horizon real-world video benchmark. 
As illustrated in Figure~\ref{fig:rebut:vlm_eval}, our framework demonstrates strong model scalability. Both Qwen2.5-VL-32B and Qwen3-VL-8B achieve competitive performance, effectively steering the robot through complex tasks. Conversely, the significant performance drop observed with the smaller Qwen2.5-VL-7B model validates the discriminative nature of our benchmark. 

\subsection{VLM Prompt}
\label{supp:vlm_prompt}


To ensure reproducibility, we include the exact prompt templates we used for the VLM. The interaction is divided into two stages: the Initial Planning Phase ($t=0$) and the Closed-Loop Monitoring Phase ($t>0$). At $t=0$, upon receiving the first observation, the system applies the \textit{Initial Prompt} to specify the task and generate the first subtask. For all subsequent steps ($t>0$), the system transitions to the \textit{Follow-up Prompt}, where the VLM uses the latest observation to assess progress on the previous instruction and produce the next one. The full templates for both phases are provided in Table~\ref{tab:prompts}.

\subsection{Demo Videos}
\label{supp:demo_video}

To provide a more intuitive understanding of our system's real-world robustness, we include three video demonstrations corresponding to the Out-of-Distribution (OOD) scenarios as we mentioned in Section 4.3 of the main paper. These demos showcase how our \method framework successfully generalizes to unseen scenarios where baseline methods typically fail.

\begin{itemize}
    \item \textbf{Demo 1: Complex Compositions.} 
    The robot is tasked with picking multiple vegetables sequentially, despite being trained only on single-object manipulation. 
    
    \item \textbf{Demo 2: Novel Objects.} 
    The robot is tasked with picking fruits and placing them on a plate, whereas the training data only contained vegetable manipulation. 

    \item \textbf{Demo 3: New Configurations.} 
    The robot performs the Clean\_Rubb task under unseen spatial layouts and background settings. This highlights the planner's robustness to novel environmental conditions and object positions.
\end{itemize}

All videos are available at \href{https://github.com/mit-han-lab/foreact}{https://github.com/mit-han-lab/foreact}.

\end{document}